\documentclass[letterpaper, 10 pt, conference]{ieeeconf}  

\IEEEoverridecommandlockouts                              

\usepackage{graphicx} 
\usepackage{amsmath} 
\usepackage{comment}
\usepackage{xcolor}
\usepackage{booktabs}
\usepackage{amssymb}
\usepackage{float}
\usepackage{multirow}
\usepackage[demo,abs]{overpic}
\usepackage{siunitx}
\usepackage{subcaption}
\usepackage{capt-of}
\makeatletter
\let\NAT@parse\undefined
\makeatother
\usepackage[bookmarks=true, colorlinks=true, citecolor=blue, linkcolor=blue]{hyperref}
\usepackage{cleveref}
\usepackage{url}
\usepackage[font=footnotesize]{caption}
\usepackage[nolist]{acronym}
\usepackage{tikz}
\usetikzlibrary{shapes,arrows,positioning,decorations.markings,calc,backgrounds,patterns}
\usepackage{pgfplots}
\pgfplotsset{compat=1.16}
\usepackage{colortbl}%
  \newcommand{\myrowcolour}{\rowcolor[gray]{0.95}}
\usepackage{placeins} 
\definecolor{ethblue}{RGB}{54,122,199}
\hypersetup{
  colorlinks,
  citecolor=ethblue,
  linkcolor=ethblue,
  urlcolor=black}
\usepackage{cite}

\title{\LARGE
ExT: Towards Scalable Autonomous Excavation via \\Large-Scale Multi-Task Pretraining and Fine-Tuning
}

\author{
Yifan Zhai$^{1,\dagger}$, Lorenzo Terenzi$^{1,\dagger}$, Patrick Frey, Diego Garcia Soto$^{1}$, Pascal Egli$^{1}$, Marco Hutter$^{1}$ \\
\thanks{$^{1}$Robotic Systems Lab, ETH Z\"urich.}%
\thanks{$^{\dagger}$These authors contributed equally to this work.}%
\thanks{This work is supported by the NCCR digital fabrication \& robotics, the SNF project No. 188596.}%
\thanks{Corresponding author: Lorenzo Terenzi, \href{mailto:lterenzi@ethz.ch}{\texttt{lterenzi@ethz.ch}}}
}

\begin{document}

\bstctlcite{IEEEexample:BSTcontrol}
\maketitle
\thispagestyle{empty}
\pagestyle{empty}

\begin{acronym}[ICANN]
\acro  {mpc}   [MPC]   {model predictive control}
\acro  {rl}    [RL]    {reinforcement learning}
\acro {ik}     [IK]    {inverse kinematics}
\acro {dt}     [DT]    {Decision Transformer}
\acro {il}     [IL]    {imitation learning}
\acro {sft}    [SFT]   {supervised fine-tuning}
\acro {rlft}   [RLFT]  {reinforcement learning fine-tuning}
\acro {rnn}    [RNN]   {recurrent neural network}
\acro {mlp}    [MLP]   {multi-layer perceptron}
\acro {ppo}    [PPO]   {proximal policy optimization}
\acro {lr}     [LR]    {learning rate}
\end{acronym}

\begin{abstract}


Scaling up the deployment of autonomous excavators is of great economic and societal importance.
Yet it remains a challenging problem, as effective systems must robustly handle unseen worksite conditions and new hardware configurations.
Current state-of-the-art approaches rely on highly engineered, task-specific controllers, which require extensive manual tuning for each new scenario. In contrast, recent advances in large-scale pretrained models have shown remarkable adaptability across tasks and embodiments in domains such as manipulation and navigation, but their applicability to heavy construction machinery remains largely unexplored.
In this work, we introduce ExT, a unified open-source framework for large-scale demonstration collection, pretraining, and fine-tuning of multi-task excavation policies. ExT policies are first trained on large-scale demonstrations collected from a mix of experts, then fine-tuned either with \ac{sft} or \ac{rlft} to specialize to new tasks or operating conditions.
Through both simulation and real-world experiments, we show that pretrained ExT policies can execute complete excavation cycles with centimeter-level accuracy, successfully transferring from simulation to real machine with performance comparable to specialized single-task controllers. Furthermore, in simulation, we demonstrate that ExT’s fine-tuning pipelines allow rapid adaptation to new tasks, out-of-distribution conditions, and machine configurations, while maintaining strong performance on previously learned tasks. These results highlight the potential of ExT to serve as a foundation for scalable and generalizable autonomous excavation.
\end{abstract}
\section{Introduction}

\begin{figure}[t]
\centering
\includegraphics[width=\columnwidth, trim=80 80 300 0, clip]{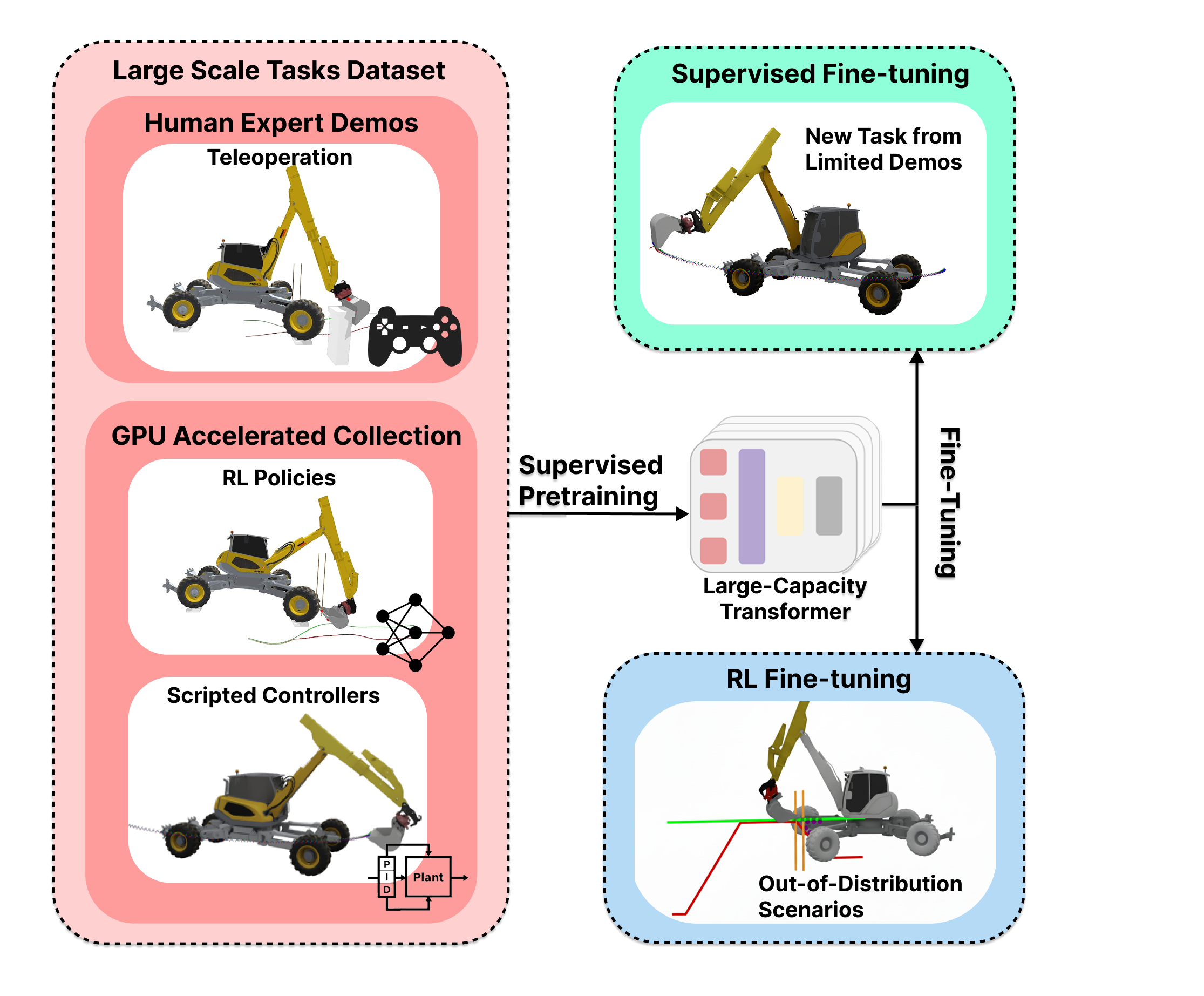}
\vspace{-0.3em}
\caption{The ExT framework enables the collection of demonstrations from different experts (human teleoperation, RL experts, scripted controllers), pretraining, followed by rapid adaptation via SFT for demonstration-only tasks or RLFT for reward-based tasks.}
\label{fig:overview}
\end{figure}

Automation of excavators promises to address workforce shortages, enhance workplace safety, and improve productivity. To achieve scalable deployment, however, autonomous excavation systems must be capable of quickly adapting to diverse operating conditions, including variations in soil properties, machine configurations, and task specifications.

While significant progress has been made in automating excavation, state-of-the-art performance is still confined to single-task, single-embodiment settings \cite{heap,jud_robotic_2021,towards_auto_ex,pascal-general-approach}. Prior work has explored adaptive digging \cite{pascal-new-rl} and low-level control \cite{nan2024adaptive}, yet these approaches do not enable task-level generalization, meaning that each new task or excavator model requires labor-intensive tuning or retraining.

Recently, large-scale pretraining of high-capacity models followed by fine-tuning has emerged as a promising paradigm \cite{openVLA, conrft, vla_rl, flare}. Pretraining on diverse demonstrations enables the model to learn useful representations and priors, while subsequent sample-efficient supervised fine-tuning (SFT) or reinforcement learning fine-tuning (RLFT) allows rapid adaptation to new tasks and embodiments.

However, compared to manipulation and navigation domains where this paradigm has been successfully applied, collecting human demonstrations on excavators is considerably more expensive. Moreover, since different approaches have been used for single-task acquisition, a unified observation modality is required for a multi-task dataset. Finally, unlike manipulation and navigation tasks that are typically evaluated solely on task completion, excavation tasks also demand consideration of safety and precision metrics, making high-quality demonstrations essential.

In this paper, we address this gap by introducing ExT, a unified framework for demonstration collection, large-scale pretraining, and rapid adaptation via fine-tuning, specifically designed for excavation tasks. 
The framework defines four core tasks that together encompass a complete excavation workflow. Using GPU-parallelized simulation, our framework can efficiently collect large numbers of demonstrations: RL agents for contact-rich tasks, scripted controller experts for high-precision tasks, and human teleoperation for additional coverage. All demonstrations share a unified observation-action interface, which is used to construct a multi-task dataset for pretraining a GPT-style transformer model. Finally, the ExT framework provides two fine-tuning methods, SFT and RLFT, for rapid skill acquisition and adaptation to new task configurations.

Through both simulated and on-machine experiments, we show that pretrained ExT policies can execute the complete excavation workflow with centimeter-level precision, achieving successful zero-shot transfer from simulation to real hardware. Our choice of a transformer architecture is validated by superior performance compared to capacity-matched RNN baselines. In simulation, we further demonstrate that ExT’s fine-tuning pipelines enable effective skill acquisition from limited demonstrations via SFT, as well as sample-efficient adaptation to unseen scenarios via RLFT, all with minimal forgetting of previously learned tasks.

Our main contributions are:
\begin{enumerate}
    \item A unified, open-source\footnote{To be released upon acceptance.} framework for excavator automation that enables large-scale demonstration collection using diverse experts, and the pretraining and fine-tuning of high-capacity transformer policies.
    \item Through simulated and on-machine experiments, we demonstrate the effectiveness of multi-task policies trained in ExT for automating excavation tasks and show that policy performance on hardware aligns with performance in simulation.
    \item Through simulated experiments, we validate the adaptation pipeline with SFT for few-shot learning and RLFT for sample-efficient adaptation, both with minimal forgetting of learned tasks.
\end{enumerate}

\section{Related Work}
\subsection{Autonomous Excavator Arm Control}
For excavator arm control in free space without external forces, classical \ac{ik} and \ac{mpc} approaches have demonstrated accurate tracking while respecting system constraints \cite{heap,traj-opt,jelavic_lstp_2023}. However, these methods are typically tailored to a specific excavator model and require retuning for different machines. A recent \ac{rl} approach proposes a more general method by incorporating data-driven actuator models of the arm joints during \ac{rl} training, achieving higher precision~\cite{pascal-general-approach}; yet, applying such methods to a new machine still necessitates recollecting joint dynamics data and retraining the policy from scratch. When external forces are present, such as during dynamic object throwing or digging, \ac{rl} methods have shown superior robustness to the sim-to-real gap and adaptability to varying operating conditions through domain randomization \cite{werner2024throw,pascal-soil-adapt}. Nevertheless, these approaches typically excel in a single task, and their performance under out-of-distribution conditions, as well as strategies for rapid adaptation, remain largely unexplored. Consequently, for full-scale, continuous excavation workflows \cite{towards_auto_ex, spinelli2025mh}, despite impressive results, existing methods face challenges in large-scale deployment, since each controller responsible for a segment of the workflow would require re-tuning or retraining when applied to new excavator models or worksites.

\subsection{Large-Scale Multi-Task Supervised Pretraining}



Large-scale pretraining via \ac{il} has achieved great success in acquiring generalist policies in domains such as manipulation~\cite{rt1, rt2, pi_0, rt-x, openVLA} and navigation~\cite{vint, nomad, habitatweb}. These approaches achieve scalability by decoupling data collection from consumption and employing large-capacity transformers that scale favorably with model size and the amount of data \cite{transformer-scaling}. By learning robust representations and versatile behavior priors, these methods demonstrate adaptability to open-world environments~\cite{pretrain} and robustness to unseen embodiments~\cite{rt-x}.

Despite these advances, high-quality demonstrations that are crucial for high-performing \ac{il} policies remain expensive to collect \cite{pirlnav}. For excavators, transformer policies trained on real-world demonstrations are constrained to the small data regime \cite{exact}. This limitation motivates the use of GPU-parallelized simulation for scalable data collection \cite{makoviychuk2021isaacgym_neuripsdb}, combining demonstrations from multiple experts, such as trained \ac{rl} policies and scripted controllers, to broaden coverage and improve policy performance.

\subsection{Fine-Tuning and Adaptation for Embodied Policies}
A key advantage of multi-task pretrained models is their adaptability to new tasks and embodiments through supervised fine-tuning (SFT) or reinforcement learning fine-tuning (RLFT). \ac{sft} uses the same behavior cloning objective from pretraining on a smaller task-specific dataset, and has proven stable and effective with only modest amounts of additional demonstrations \cite{rt2, openVLA, vpt}. \ac{rlft} leverages the robust representations, versatile behavior priors, and good inductive biases acquired during pretraining to fine-tuning the pretrained policy with great sample-efficiency~\cite{flare}. However, directly applying \ac{rl} gradient updates to large-scale pretrained policies often leads to oscillation or collapse in performance. To address this, several techniques have been proposed to stabilize training, including critic-only warmup for actor-critic methods \cite{pirlnav}, reduced learning rates with no entropy bonus \cite{flare}, KL-divergence regularization between the updated and pretrained policies \cite{vpt}, and alternating \ac{rlft} with \ac{sft} to mitigate catastrophic forgetting~\cite{vla_rl}. These methods have unlocked substantial performance improvements on top of large-scale pretraining, achieving state-of-the-art results across diverse domains \cite{conrft, flare, vla_rl}.


\section{Method}
\label{sec:method}

In this section, we detail the implementation of the proposed ExT framework. ExT consists of three main stages. In the first stage, we define four excavation tasks in simulation (Section~\ref{sec:task-def}) and generate large-scale demonstrations (Section~\ref{sec:data}) using GPU-parallelized simulations in IsaacGym~\cite{makoviychuk2021isaacgym_neuripsdb}. These demonstrations combine trajectories from RL experts for contact-rich tasks, scripted controllers for precise movements, and a limited number of human teleoperation demonstrations when neither is available. In the second stage, ExT pretrains high-capacity models on this multi-task dataset to learn versatile representations (Section~\ref{sec:ExT-model}). In the final stage, ExT adapts policies through \ac{sft} for rapid skill acquisition with small datasets and KL-regularized PPO (\ac{rlft}) for on-policy adaptation to out-of-distribution scenarios (Section~\ref{sec:adaptation}).

\subsection{Task Definitions}
\label{sec:task-def}

\begin{figure}[!t]
\vspace*{12pt} 
\centering
\definecolor{digcolor}{RGB}{70,130,200}
\definecolor{dumpcolor}{RGB}{255,140,0}
\definecolor{movecolor}{RGB}{50,180,50}
\definecolor{abortcolor}{RGB}{200,50,50}
\vspace{-0.2em}
\noindent\fcolorbox{digcolor}{white}{%
\begin{minipage}{\dimexpr\linewidth-2\fboxsep-2\fboxrule}
\vspace{0.5em}
{\centering\scriptsize\textbf{DIG TASK}\\}
\begin{subfigure}{.47\linewidth}
  \centering
  \includegraphics[width=\linewidth, trim={13cm 16.5cm 10cm  13cm},clip]{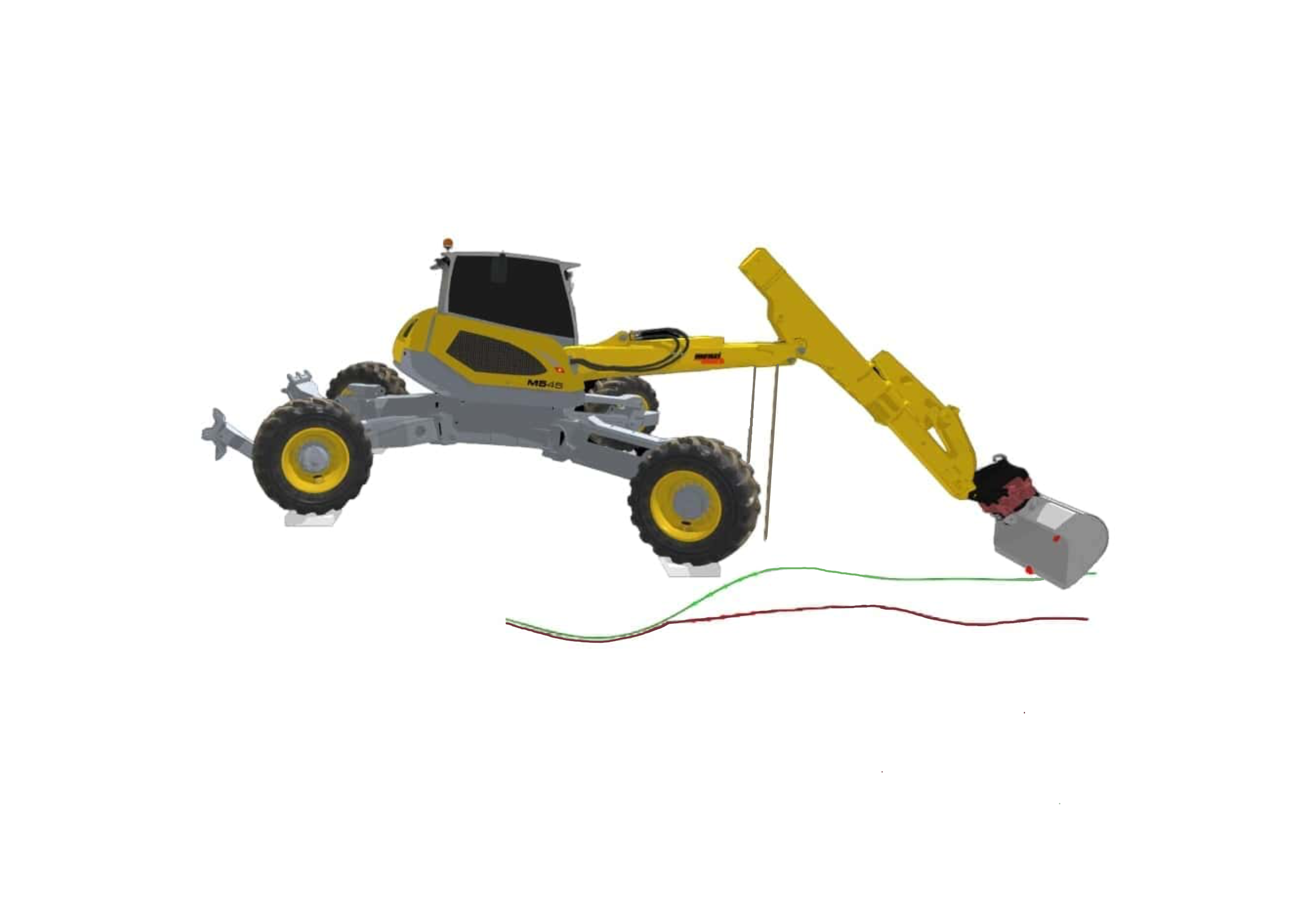}
  \caption{}
  \label{fig:dig-start}
\end{subfigure}%
\hfill
\begin{subfigure}{.47\linewidth}
  \centering
  \includegraphics[width=\linewidth, trim={15cm 9.5cm 8cm  13cm},clip]{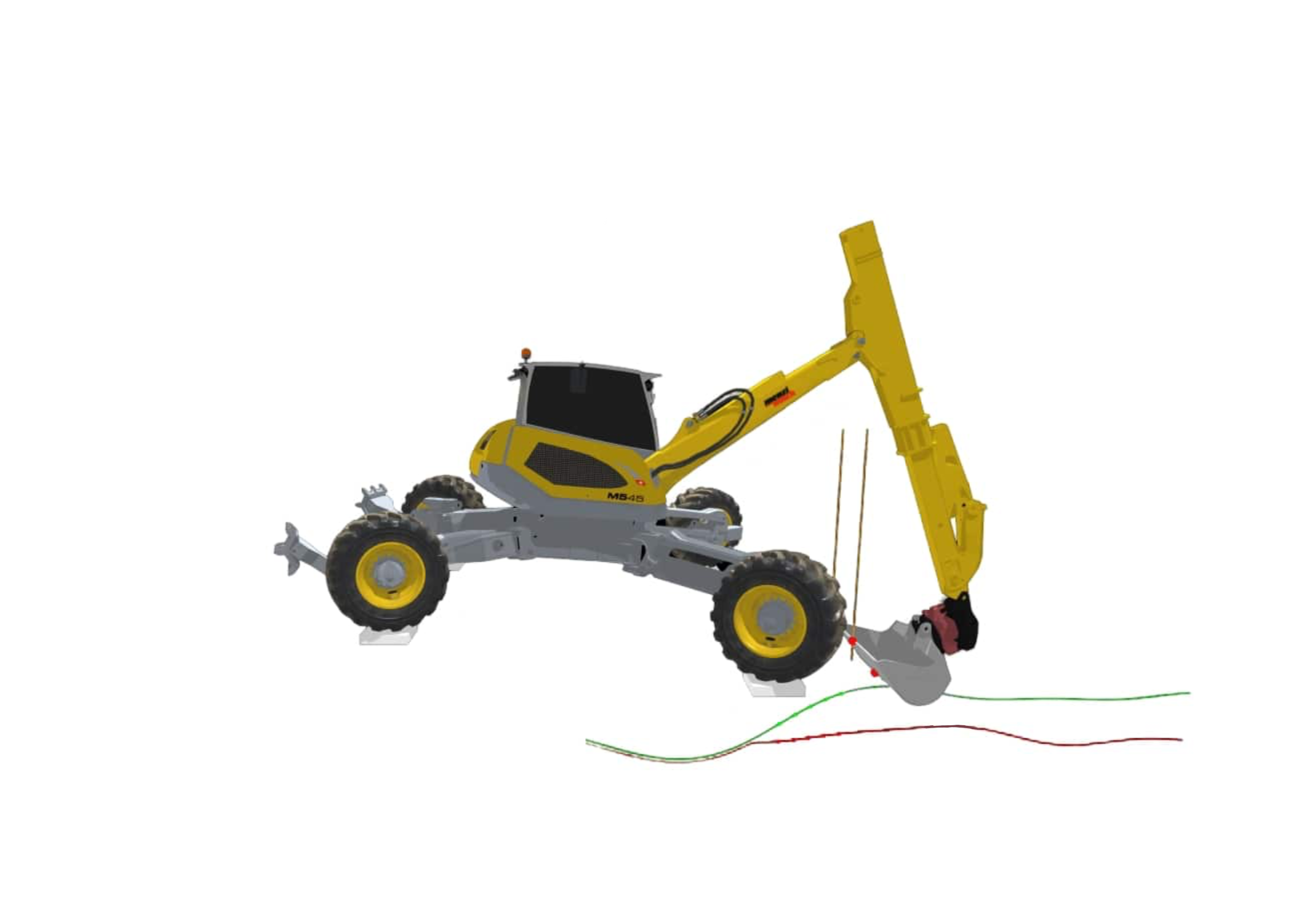}
  \caption{}
  \label{fig:dig-end}
\end{subfigure}
\end{minipage}}

\vspace{0.15em}

\noindent\fcolorbox{abortcolor}{white}{%
\begin{minipage}{\dimexpr\linewidth-2\fboxsep-2\fboxrule}
\vspace{0.5em}
{\centering\scriptsize\textbf{ABORT DIGGING \& RESET}\\[0.3em]}
\begin{subfigure}{.24\linewidth}
  \centering
  \includegraphics[trim={35cm 7cm 10cm  7cm},clip, width=\linewidth]{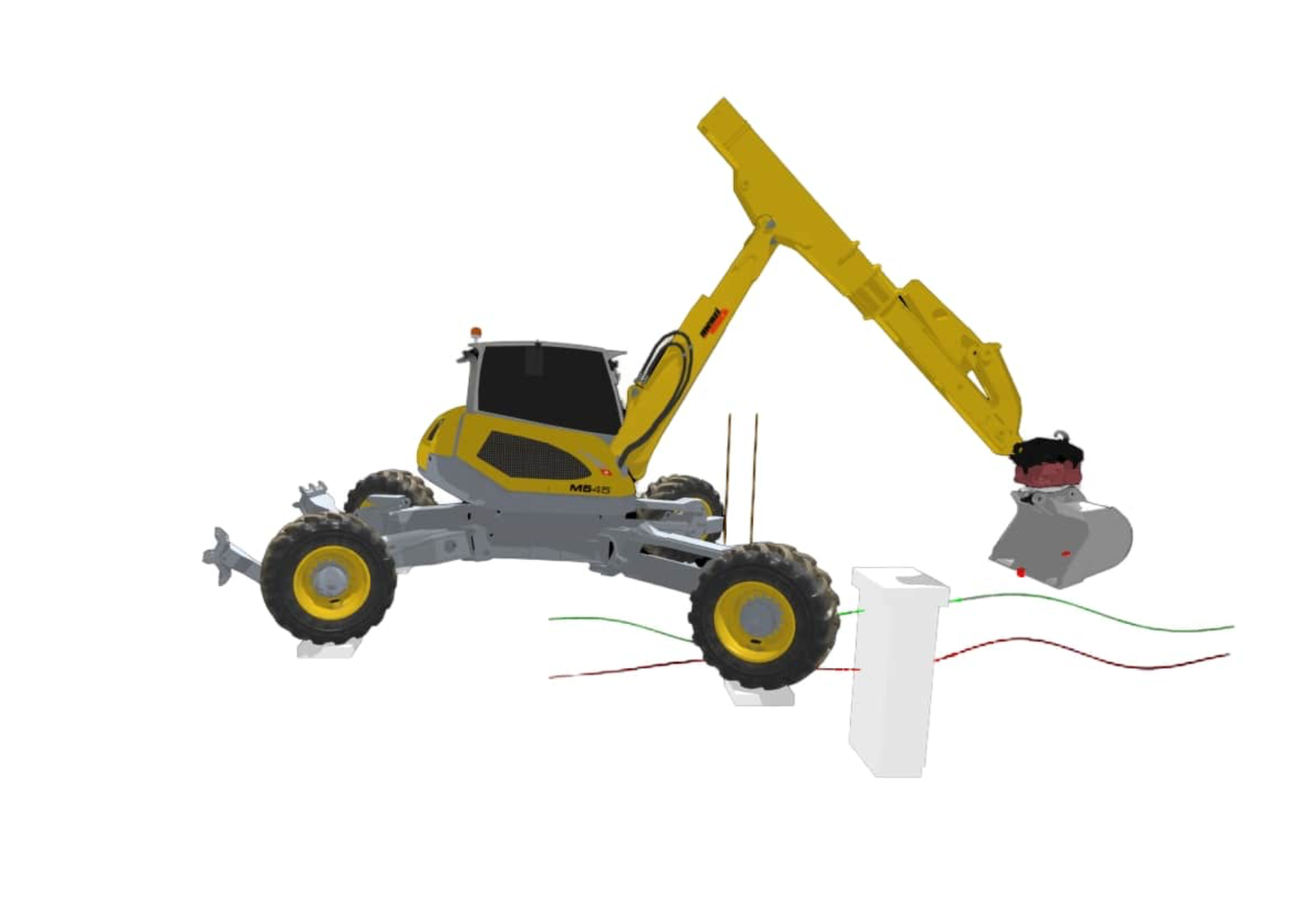}
  \caption{}
  \label{fig:obst-start}
\end{subfigure}%
\hfill
\begin{subfigure}{.24\linewidth}
  \centering
  \includegraphics[trim={31cm 7cm 16cm 5cm},clip, width=\linewidth]{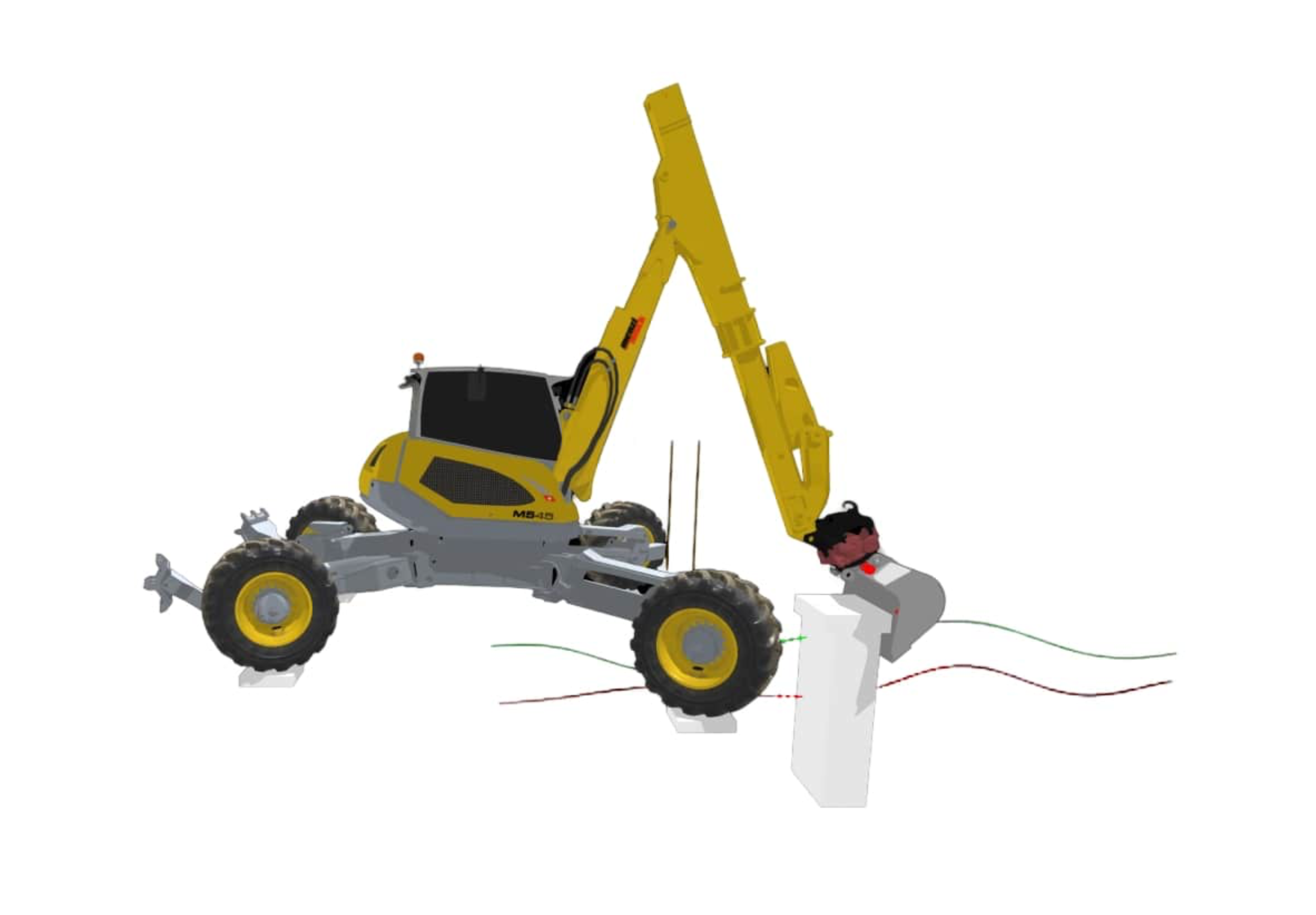}
  \caption{}
  \label{fig:obst-stuck}
\end{subfigure}%
\hfill
\begin{subfigure}{.24\linewidth}
  \centering
  \includegraphics[trim={31cm 7cm 16cm 5cm},clip, width=\linewidth]{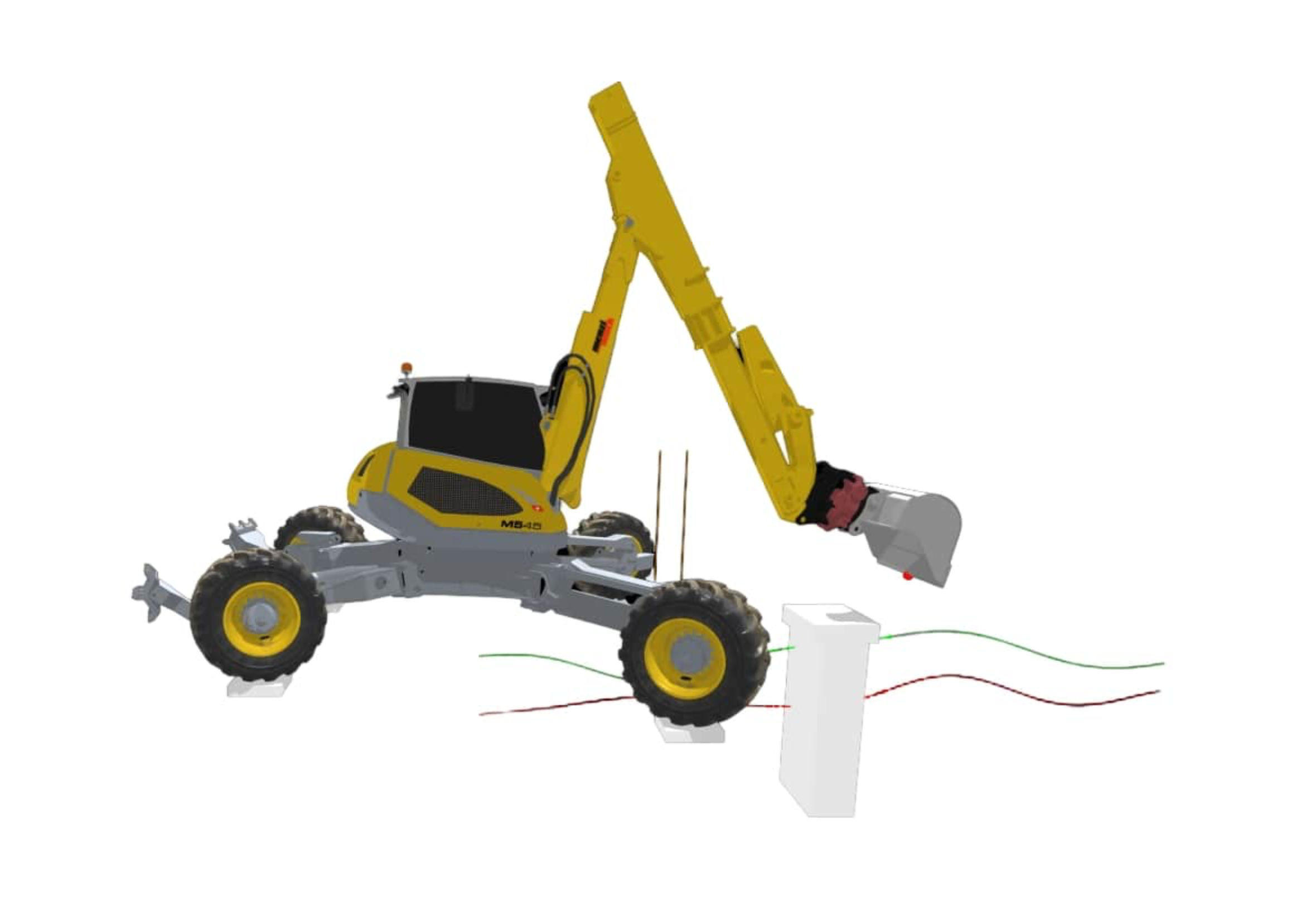}
  \caption{}
  \label{fig:obst-recover}
\end{subfigure}%
\hfill
\begin{subfigure}{.24\linewidth}
  \centering
  \includegraphics[trim={31cm 7cm 16cm 5cm},clip, width=\linewidth]{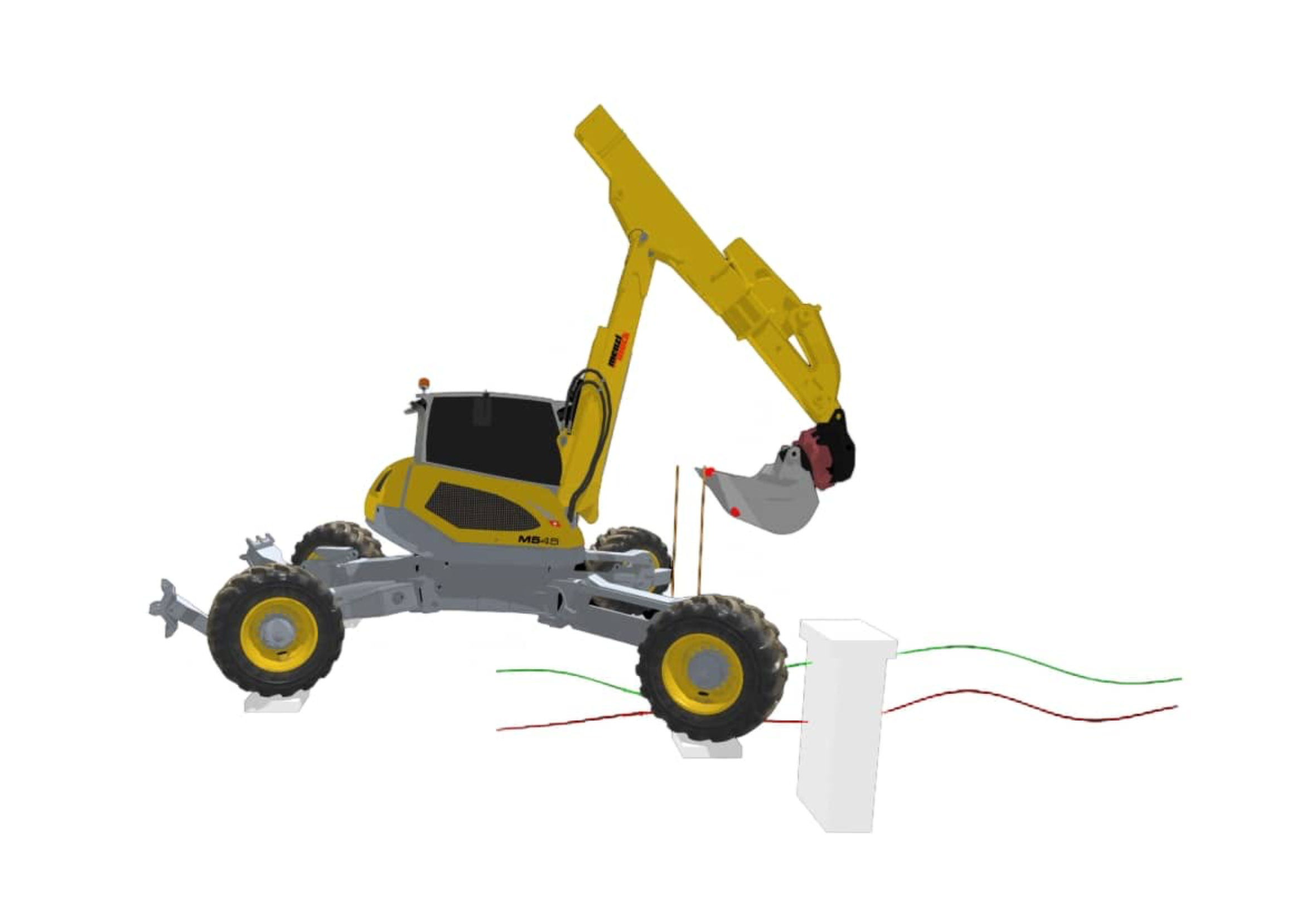}
  \caption{}
  \label{fig:obst-end}
\end{subfigure}
\end{minipage}}

\vspace{0.15em}

\noindent\fcolorbox{dumpcolor}{white}{%
\begin{minipage}{\dimexpr\linewidth-2\fboxsep-2\fboxrule}
\vspace{0.5em}
{\centering\scriptsize\textbf{DUMP TASK}\\}
\begin{subfigure}{.49\linewidth}
  \centering
  \includegraphics[trim={7cm 6cm 10cm 5cm},clip, width=\linewidth]{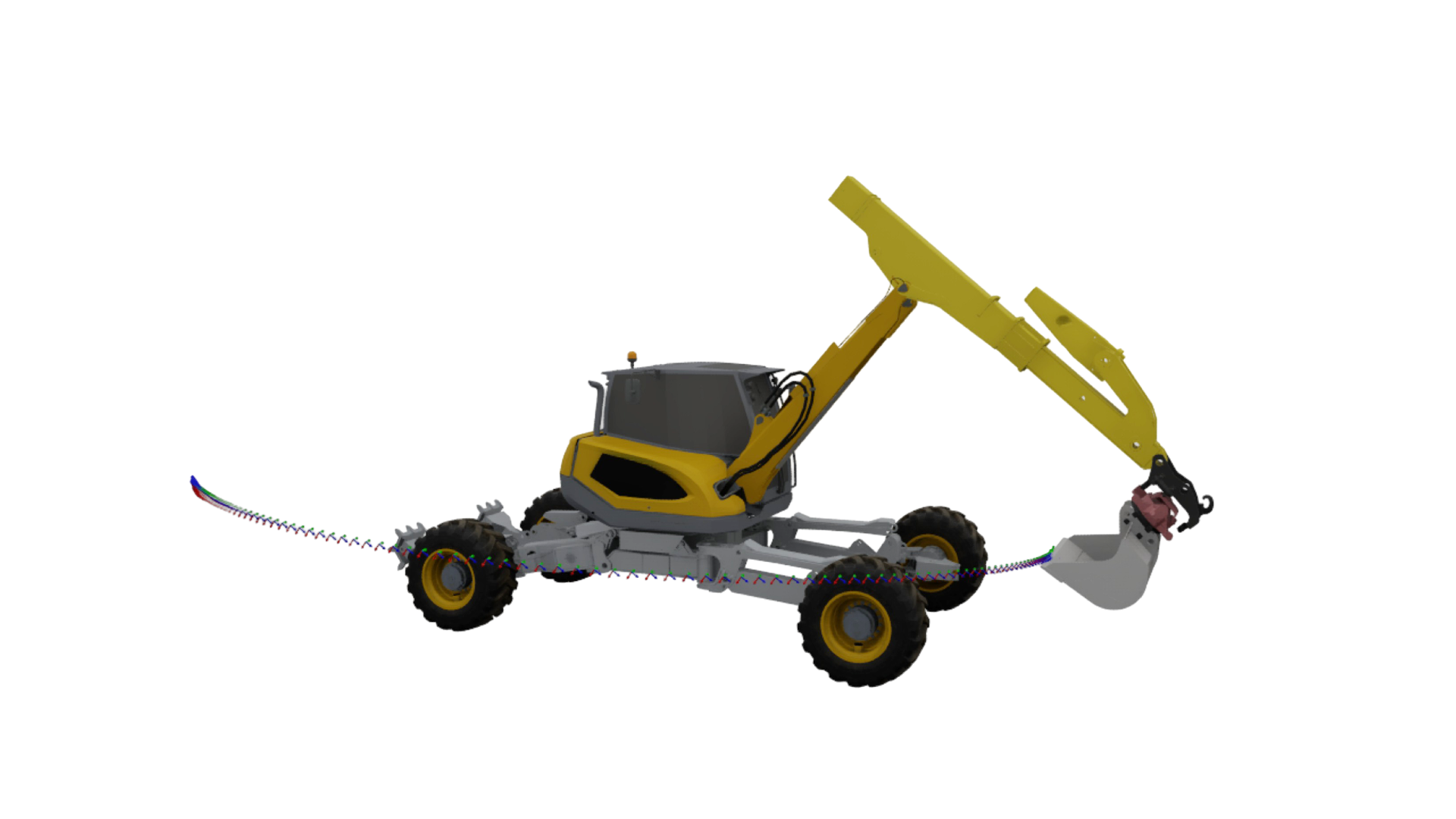}
  \caption{}
  \label{fig:dump-start}
\end{subfigure}%
\hfill
\begin{subfigure}{.45\linewidth}
  \centering
  \includegraphics[trim={10cm 7cm 20cm 5cm},clip, width=\linewidth]{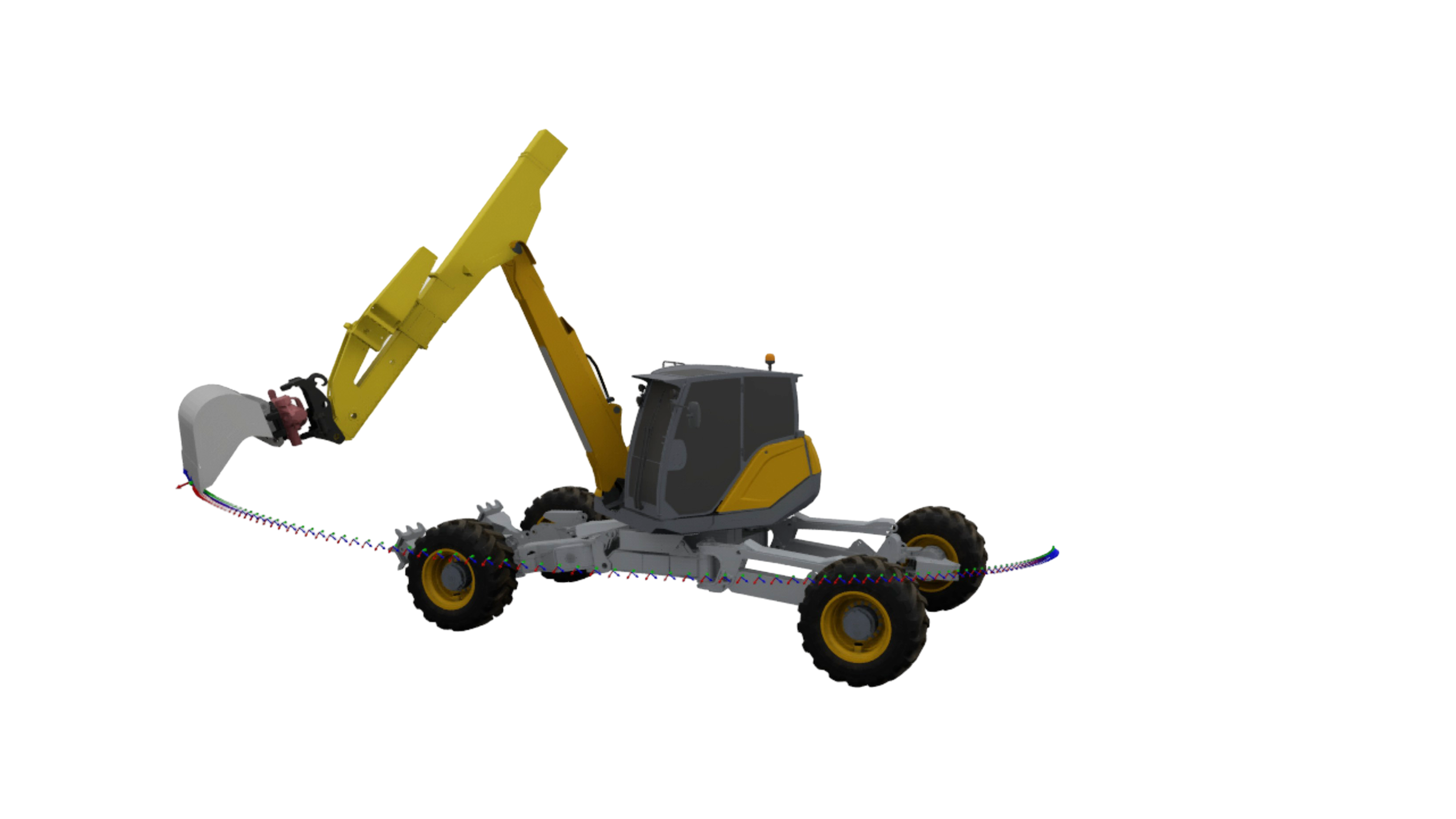}
  \caption{}
  \label{fig:dump-end}
\end{subfigure}
\end{minipage}}

\vspace{0.15em}

\noindent\fcolorbox{movecolor}{white}{%
\begin{minipage}{\dimexpr\linewidth-2\fboxsep-2\fboxrule}
\vspace{0.5em}
{\centering\scriptsize\textbf{MOVE ARM TASK}\\[0.3em]}
\begin{subfigure}{.49\linewidth}
  \centering
  \includegraphics[trim={0cm 5.5cm 15cm 5cm},clip,width=\linewidth]{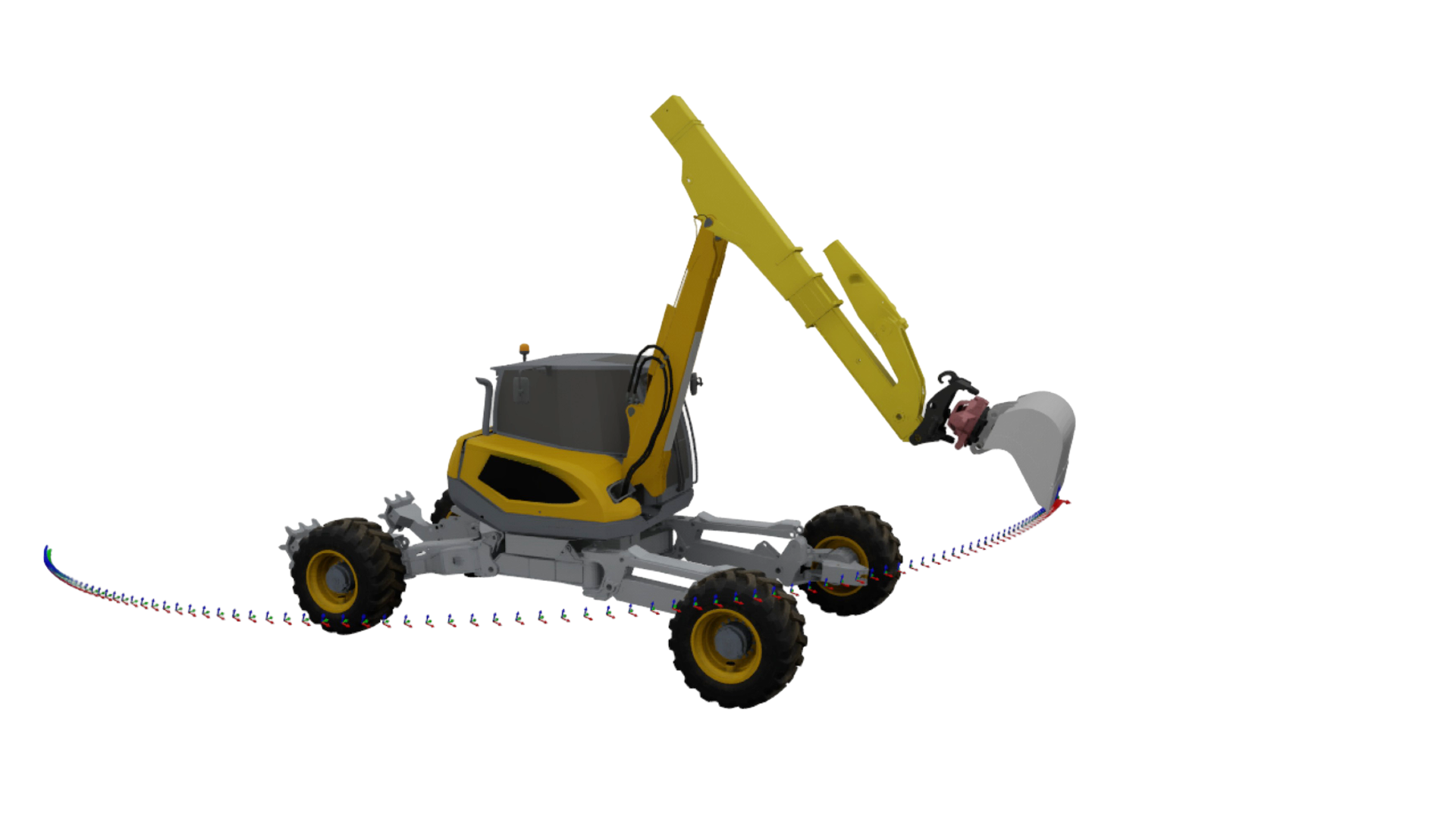}
  \caption{}
  \label{fig:find-dig-start}
\end{subfigure}%
\hfill
\begin{subfigure}{.49\linewidth}
  \centering
  \includegraphics[trim={5cm 5.5cm 10cm 5cm},clip,width=\linewidth]{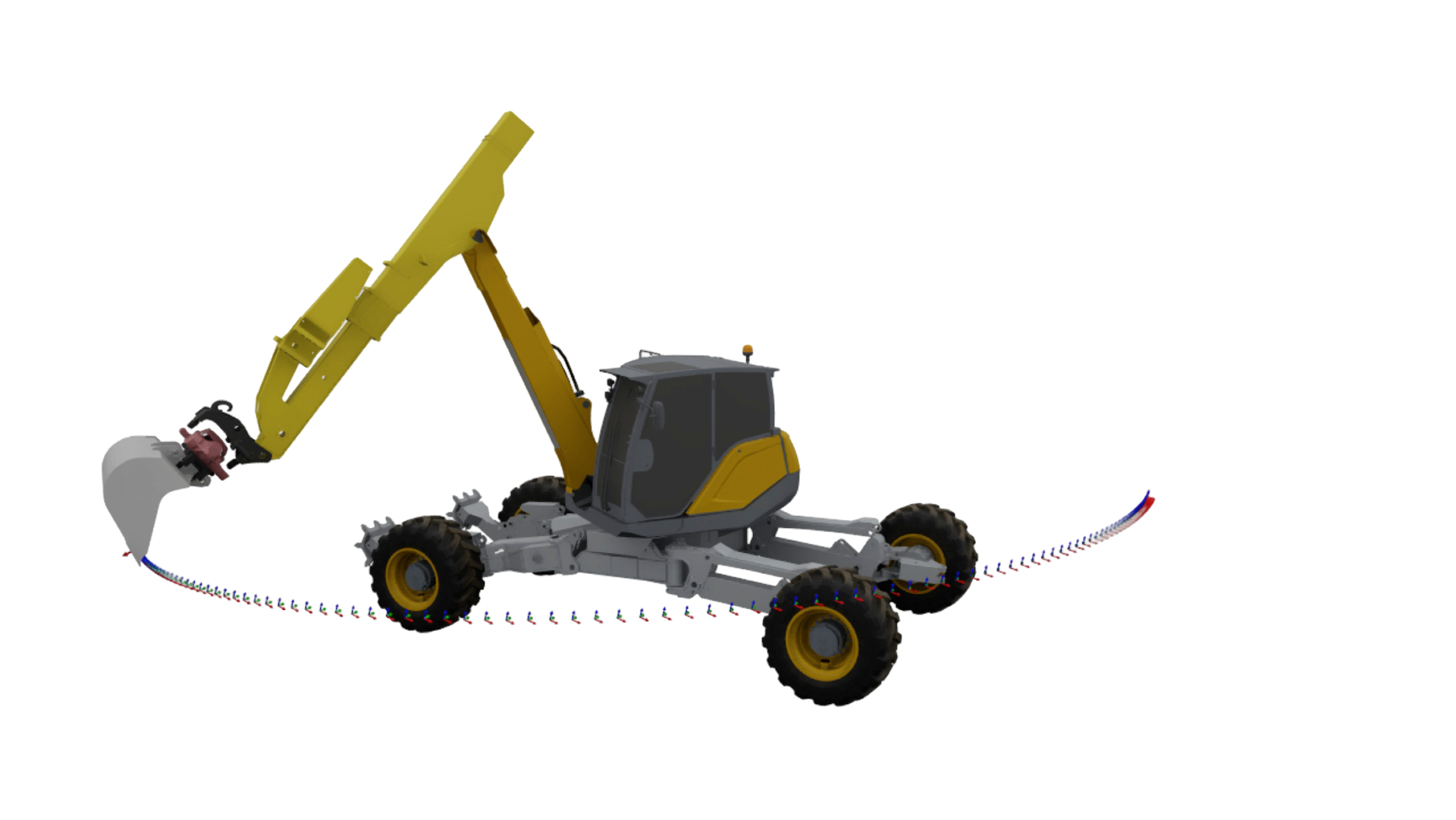}
  \caption{}
  \label{fig:find-dig-end}
\end{subfigure}
\end{minipage}}

\caption{Task Visualizations. \textbf{Dig}: (a) initial and (b) final state, with soil profile (green), max digging depth (red), and pull-up line (brown) visualized. \textbf{Abort Digging \& Reset}: (c) normal digging, (d) collision with obstacle (white), (e)~recovery, and (f) retraction. \textbf{Dump}: (g)~initial pose with full shovel and (h)~move to target and open shovel, with shovel trajectory visualized. \textbf{Move Arm}: move open shovel from (i)~initial pose to (j)~target pose.}
\label{fig:excavation-tasks}
\vspace{-0.5em}
\end{figure}

The four tasks, as shown in Fig.~\ref{fig:excavation-tasks}, together form a complete dig--dump--move excavation cycle:
\begin{itemize}
    \item \textbf{Dig}: Following the formulation in \cite{pascal-new-rl}, the policy must fill the bucket under diverse soil types (modeled with the fundamental earth moving equations), soil profiles, and initial arm and base configurations, while satisfying task constraints (e.g., maximum depth, pull-up distance).
    \item \textbf{Abort Digging \& Reset}: If, during the \emph{dig} task, the bucket collides with a hidden, immovable object (e.g., rock, pipe), the policy must detect the stall from proprioceptive observations, recover by opening the shovel, and retract to a safe pose. The policy has no prior observation or knowledge of the obstacle.
    \item \textbf{Dump}: The policy moves the full shovel to a designated location and releases the material by opening the shovel.
    \item \textbf{Move Arm}: The policy moves the open shovel to a designated location for the next digging cycle.
\end{itemize}

\subsection{Large-Scale Demonstration Generation}
\label{sec:data}



\begin{table}[t]
\vspace*{8pt} 
\begin{center}
\caption{Shared state-action interface for all tasks}
\label{dt-states}
\setlength\tabcolsep{0.8ex}
\begin{tabular}{ l c | c c c c} 
\toprule
\textbf{State} $s_t$ & \textbf{Dim.}  &  \emph{dig}   & \shortstack{\emph{abort} \& \\ \emph{reset}} & \emph{dump}  & \shortstack{\emph{move} \\ \emph{arm}}\\
\hline
 \myrowcolour
 Task encoding                      & 3        & \checkmark & \checkmark & \checkmark & \checkmark\\
 Task target                        & 9        &  &  & \checkmark & \checkmark\\ \myrowcolour
 Cabin turn joint pos. \& vel       & 2        &  &  & \checkmark & \checkmark\\ 
 Arm joint pos. \& vel.             & 8        & \checkmark & \checkmark & \checkmark & \checkmark\\ \myrowcolour
 Arm joint torque                   & 4        & \checkmark & \checkmark &            & \\
 Previous arm joint action          & 4        & \checkmark & \checkmark &            & \\ \myrowcolour
 Bucket linear pos. \& vel.         & 4        & \checkmark & \checkmark &            & \\
 Bucket angular pos. \& vel.        & 2        & \checkmark & \checkmark &            & \\ \myrowcolour
 Bucket linear vel. norm            & 1        & \checkmark & \checkmark &            & \\
 Base pitch pos. \& vel.            & 2        & \checkmark & \checkmark &            & \\ \myrowcolour
 Bucket fill ratio                  & 1        & \checkmark & \checkmark &            & \\
 Bucket angle of attack             & 1        & \checkmark & \checkmark &            & \\ \myrowcolour
 Soil height in base frame          & 5        & \checkmark & \checkmark &            & \\
 Bucket depth in soil               & 1        & \checkmark & \checkmark &            & \\ \myrowcolour
 Bucket pitch joint vel.            & 2        & \checkmark & \checkmark &            & \\
 Max digging depth                  & 5        & \checkmark & \checkmark &            & \\ \myrowcolour
 Pull-up distance                   & 1        & \checkmark & \checkmark &            & \\
 \hline
 & 55 \\
\toprule
\textbf{Action} $a_t$ & \\
\hline
\myrowcolour
Cabin joint velocity & 1 &           &            & \checkmark & \checkmark\\
Arm joint velocities & 4 & \checkmark & \checkmark & \checkmark & \checkmark\\
\toprule
\end{tabular}
\end{center}
\vspace{-0.2cm}
\end{table}

ExT relies on GPU-parallelized simulation (IsaacGym~\cite{makoviychuk2021isaacgym_neuripsdb,mittal2023orbit}) to generate large-scale, high-quality demonstrations using a mix of experts. All demonstrations share a unified observation-action interface, as shown in Table~\ref{dt-states}. The experts used and their corresponding tasks are as follows:

\textbf{\textit{\ac{rl} Experts} for Dig task.} \ac{rl} methods have been shown to excel in contact-rich tasks \cite{werner2024throw, pascal-soil-adapt}. In the case of excavation, digging requires adapting to complex soil--bucket interactions and leveraging proprioceptive feedback (joint torques) to handle resistance forces. Following \cite{pascal-new-rl}, we apply domain randomization over soil types, soil profiles, maximum digging depths, and initial excavator configurations to train an \ac{rl} expert, which achieves a success rate of approximately $98\%$. The same domain randomization is employed when collecting demonstrations with the trained expert, yielding 150{,}000 successful episodes, equivalent to about 15 days of continuous real-world digging. Notably, reproducing such a wide range of domain variations in the real world would be infeasible, highlighting the advantage of large-scale simulation for data generation.

\textbf{\textit{Scripted Policies} for Dump and Move Arm tasks.} For free-space motions with high precision requirements and no external forces, we employ a GPU-parallelized trajectory planner coupled with a PID tracking controller. Start and target shovel positions are randomly sampled within the excavator’s reachable workspace to ensure comprehensive coverage, and 150{,}000 episodes are collected for each task. This dataset, corresponding to roughly 30 days of continuous real-world operation, is generated in under two hours on a single RTX~3090 GPU.

\textbf{\textit{Human Teleoperation} for Abort \& Reset task.} To account for abnormal operation scenarios, such as collisions or stalling during digging, we randomly place obstacles in the digging environment. The PPO expert performs digging until a collision occurs, at which point a human operator takes over via teleoperation to execute the recovery phase. We collect 2{,}000 episodes in this manner, requiring approximately three hours of wall-clock time. Although smaller, this dataset is valuable because it avoids potential machine damage and enables randomization of obstacle locations that would be difficult to achieve in the real world.

\subsection{Pretraining and Model Architectures}
\label{sec:ExT-model}

We follow the standard behavior cloning setup for pretraining. Given the expert dataset $D = \{(s_1, a_1), (s_2, a_2), \dots, (s_i, a_i)\}$, we minimize the L1 loss on the predicted actions:
\begin{equation*}
    J^{BC}(\theta) = \mathbb{E}_{(s, a) \sim D} \big[ |\pi_\theta(s) - a| \big].
\label{eqn:l1}
\end{equation*}
After pretraining, we obtain the pretrained model $\pi_0$.  

Following successful approaches in prior work, we adopt a GPT-style decoder-only transformer with 6 layers and 6 attention heads. At each timestep, the previous action and current state (as defined in Table~\ref{dt-states}) are embedded via linear layers with a hidden dimension of 640. Any unused state-action entries are set to zeros. The model uses a context window of $K=25$ steps, and joint velocities are predicted autoregressively during inference, resulting in a high-capacity transformer with approximately 25M parameters.  

Since transformer-based models for excavation automation are still underexplored, we also implement a sequential 5-layer LSTM baseline with 1024 hidden dimensions, scaled to match the parameter count of the GPT model, to validate the choice of architecture for pretraining.



\subsection{Fine-Tuning Pipelines}
\label{sec:adaptation}

After obtaining the pretrained policy $\pi_0$, ExT provides two fine-tuning methods for different use cases: \ac{sft} enables rapid acquisition of new skills from a small number of demonstrations, while \ac{rlft} allows adaptation to out-of-distribution scenarios when a simulation environment and reward function are available, such as new terrain profiles, soil parameters, or bucket geometries for the \emph{dig} task.

\subsubsection{Supervised Fine-Tuning (SFT)}
\label{sec:sft}

Supervised fine-tuning updates the weights of $\pi_0$ using the same L1 loss in (\ref{eqn:l1}), with the same optimizer and learning rate schedule, but applied to a much smaller dataset containing demonstrations of new tasks or out-of-distribution conditions. When deploying ExT policies in an excavation pipeline involving multiple tasks, preventing catastrophic forgetting of previously learned behaviors is crucial. To address this, we interleave a small replay buffer of pretraining demonstrations from other tasks with the new SFT dataset during fine-tuning.

\subsubsection{Reinforcement Learning Fine-Tuning (RLFT)}
\label{sec:rlft_method}
Compared to \ac{sft}, applying \ac{rlft} directly on pretrained policies often leads to instability without careful treatment \cite{flare, pirlnav, vpt, vla_rl}, manifesting as catastrophic forgetting and gradient instabilities that degrade the pretrained representations. Building on prior work, we propose a practical recipe that stabilizes \ac{ppo}-based fine-tuning through four key design choices. 

First, we use learning rates much smaller than those typical in RL \cite{flare}. While standard \ac{ppo} often uses \(10^{-3}\)–\(10^{-4}\), we set the actor learning rate to \(10^{-5}\) with cosine annealing to \(10^{-7}\), which limits disruptive updates to the pretrained policy. Second, we train a separate critic warm-started from rollouts of the pretrained policy for 100 steps \cite{flare, pirlnav}, keeping actor and critic learning balanced for stability. Third, we use a smaller entropy coefficient $c_2$ and parameterize the action log standard deviations independently of the main policy and the current state \cite{flare, rsl_ppo}, allowing limited exploration without drastically shifting the pretrained action means. Finally, we apply a KL penalty toward the frozen pretrained policy \(\pi_0\) \cite{vpt}, anchoring the fine-tuned policy to prior behaviors. The modified PPO objective is
\begin{align*}
L^{\text{PPO}}(\theta) = & \mathbb{E}_t \Big[ L^{\text{CLIP}}_t(\theta) 
- c_1 \, L^{\text{VF}}_t(\theta) 
+ c_2 \, S[\pi_\theta](s_t) \Big]\\
& -\ \beta\,\mathrm{KL}\!\left(\pi_{\theta}\,\|\,\pi_0\right)
\end{align*}
with \(c_2{=}0.0005\) and \(\beta{=}0.02\), encouraging low-variance exploration while keeping \(\pi_{\theta}\) close to \(\pi_0\).

%


During \ac{rlft}, the transformer outputs are used as mean actions, while separate state-independent log standard deviations are learned \cite{rsl_ppo}. Actions are sampled from a tanh-squashed Gaussian during training and set to the mean at test time. A separate 3-layer MLP critic (128 hidden dimensions, ELU) is warm-started for 100 steps using rollouts from the pretrained policy. We use separate Adam optimizers for the actor ($lr=10^{-5}$ annealed to $10^{-7}$), the critic ($lr=10^{-4}$), and the standard deviation parameters ($lr=5{\times}10^{-4}$).


\section{Experiments \& Results}
\label{sec:experiments}
We first define evaluation metrics for each task (Section \ref{sec:task_metrics}). Then, the experiment results are organized into pretraining (Section \ref{sec:multitask_performance}), \ac{sft} (Section \ref{sec:exp_sft}), and \ac{rlft} (Section \ref{sec:rlft}).

\subsection{Task Evaluation Metrics}
\label{sec:task_metrics}
For the \emph{dig} task, we measure success as the rate of positive terminations, with failures defined as violations of any task constraints specified in \cite{pascal-new-rl}, evaluated over 1000 random configurations. The \emph{abort digging \& reset} task is tested on 200 random configurations and consists of two phases: (i) a normal digging phase, evaluated using the same success criterion as the \emph{dig} task, and (ii) a recovery phase, where the policy must recover from collision-induced stalling and retract the arm to a safe pose, with success measured as the completion rate of this recovery. For the \emph{dump} and \emph{move arm} tasks, we evaluate 25 random start--target position pairs and report the minimum distance between the shovel and target location while maintaining end-effector velocity below \SI{0.2}{\meter\per\second}. The \emph{dump} task further requires the shovel to be fully opened for the episode to count as successful.

\newcommand{\NA}{\textemdash}
\begin{table*}[!th]
\vspace*{8pt} 
\centering
\caption{Results of Simulated Experiments Across Four Tasks Using Pretrained GPT and RNN Policies}
\label{sim_res_table}
{%
\begin{tabular}{l c | c c | c c | c c}
\toprule
 & \multicolumn{1}{c}{\textbf{Dig Task}} & \multicolumn{2}{c}{\textbf{Abort Digging \& Reset Task}} & \multicolumn{2}{c}{\textbf{Dump Task}}& \multicolumn{2}{c}{\textbf{Move Arm Task}} \\
\cmidrule(lr){2-2} \cmidrule(lr){3-4} \cmidrule(lr){5-6} \cmidrule(lr){7-8}
Architecture-Dataset & Success & \shortstack{Digging Phase \\ Success} & \shortstack{Recovery Phase \\ Success} & \shortstack{Pos. Err. (cm)} & \shortstack{Pos. Err. (cm) \\w/ Perturbation } & \shortstack{Pos. Err. (cm)} & \shortstack{Pos. Err. (cm) \\w/ Perturbation}\\
\midrule
\myrowcolour
RNN-Dig                 & 98.8\%    & \NA       & \NA       & \NA           & \NA           & \NA           & \NA       \\ 
RNN-Recover             & \NA       & 36.3\%    & 100\%     & \NA           & \NA           & \NA           & \NA       \\ 
\myrowcolour
RNN-Dig+Recover         & 97.8\%    & 99.0\%    & 99.5\%    & \NA           & \NA           & \NA           & \NA       \\ 
RNN-Dump                & \NA       & \NA       & \NA       & $3.1\pm2.2$   & $13.7\pm9.7$  & \NA           & \NA       \\ 
\myrowcolour
RNN-Move                & \NA       & \NA       & \NA       & \NA           & \NA           & $3.8\pm6.7^*$ & $8.5\pm7.0^*$\\
RNN-Dump+Move           & \NA       & \NA       & \NA       & $21.8\pm15.8$ & $40.8\pm29.3$ & $10.4\pm6.5$  & $14.7\pm9.8$\\ \myrowcolour
RNN-4-Tasks             & 98.1\%    & 99.0\%    & 99.5\%    & $21.7\pm15.5$ & $44.3\pm46.4$ & $27.8\pm27.7$ & $32.8\pm27.8$\\
\cmidrule(lr){1-8}
GPT-Dig                 & 98.6\%    & \NA       & \NA       & \NA           & \NA           & \NA           & \NA       \\
\myrowcolour
GPT-Recover             & \NA       & 59.0\%    & 99.5\%    & \NA           & \NA           & \NA           & \NA       \\ 
GPT-Dig+Recover         & 98.6\%    & 99.0\%    & 98.0\%    & \NA           & \NA           & \NA           & \NA       \\
\myrowcolour
GPT-Dump                & \NA       & \NA       & \NA       & $2.4\pm0.9$   & $8.0\pm6.5$   & \NA           & \NA       \\ 
GPT-Move                & \NA       & \NA       & \NA       & \NA           & \NA           & $0.5 \pm 0.3$ & $42.2\pm41.4$\\
\myrowcolour
GPT-Dump+Move           & \NA       & \NA       & \NA       & $\mathbf{1.0\pm0.7}$   & $\mathbf{3.3\pm3.7}$   & $\mathbf{0.5 \pm 0.5}$ & $6.7\pm7.3$\\ 
GPT-4-Tasks             & \textbf{98.8\%}   & \textbf{99.0\%}    & \textbf{99.5\%}    & $1.6\pm0.9$   & $5.5\pm3.4$   & $0.8 \pm 0.4$ & $\mathbf{5.3\pm2.5}$\\
\bottomrule
\multicolumn{8}{l}{\footnotesize $^*$ A failed episode with error $>\SI{10}{\meter}$ removed from the calculation.}
\end{tabular}
}
\end{table*}

\subsection{Pretraining Experiments}
\label{sec:multitask_performance}

To validate the large-scale pretraining procedure, we train multiple GPT-style transformers and RNN baselines on various combinations of task data and evaluate their performance to assess the effects of multi-task pretraining on individual tasks. We further deploy the pretrained policies on a Menzi Muck M545 excavator, demonstrating zero-shot sim-to-real transfer through quantitative evaluations and illustrating real-world applicability via a complete excavation workflow. 

\subsubsection{Simulated Experiments}
Table~\ref{sim_res_table} summarizes the different combinations of task data used to train the GPT policies and RNN baselines, along with their performance on each task in simulation. From these quantitative results, several key insights emerge. \textbf{Transformers are more suitable for large-scale multi-task pretraining:} GPT-style transformers maintain high performance even when trained on multiple tasks simultaneously, whereas RNNs, while capable, degrade more noticeably as the number of tasks increases. \textbf{Transformers excel at precision tasks:} For high-precision control tasks such as \emph{move arm} and \emph{dump}, GPT-based models achieve consistently accurate performance across tasks and environments, while RNNs exhibit larger errors, with single-task RNN policies occasionally failing catastrophically. This suggests that transformers better capture the representations needed for fine-grained control. \textbf{Positive transfer between related tasks:} Training on both \emph{dig} and \emph{abort \& reset} tasks improves digging phase performance compared to training on \emph{abort \& reset} alone, indicating that simultaneously learning closely related tasks enhances individual task performance. A similar trend is observed for the \emph{Dump} and \emph{Move Arm} tasks, further highlighting the advantage of positive skill transfer from multi-task training.

\begin{table}[t]
\begin{center}
\caption{On-Machine Results for \emph{Dump} and \emph{Move Arm}}
\label{ik-results-real}
\begin{tabular}{ l  c | c} 
\toprule
 & \multicolumn{1}{c}{Dump Task} & \multicolumn{1}{c}{Move Arm Task} \\
 \cmidrule(lr){2-2} \cmidrule(lr){3-3}
 Policy  & Pos. Err. (cm) & Pos. Err. (cm) \\
\midrule
\myrowcolour
    GPT-Dump                    & $8.5 \pm 4.3$ & \NA\\
    GPT-Move                    & \NA           & $11.4 \pm 7.7$\\
\myrowcolour
    GPT-Dump+Move               & $5.0 \pm 4.8$ & $2.2 \pm 1.2$ \\
    GPT-4-Tasks                 & $\mathbf{3.6 \pm 2.5}$  & $\mathbf{2.2 \pm 1.0}$\\
\bottomrule
\end{tabular}
\end{center}
\vspace{-0.3cm}
\end{table}

\subsubsection{On-Machine Deployment}
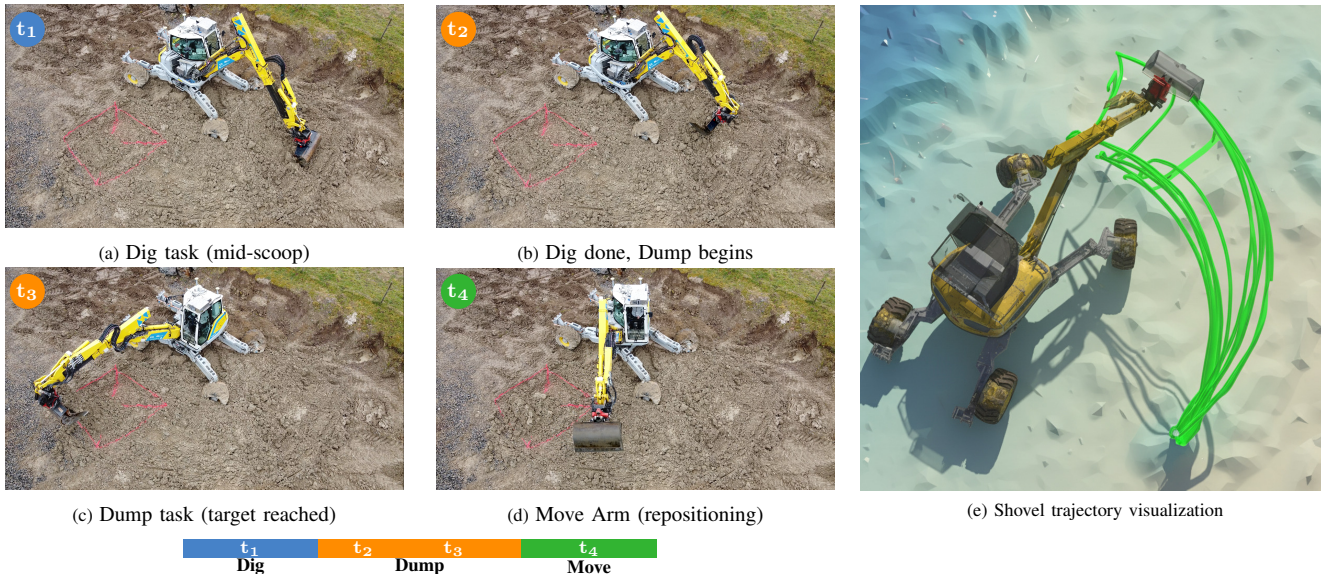
\begin{figure*}[t]
\vspace*{12pt} 
\centering
\definecolor{digcolor}{RGB}{70,130,200}
\definecolor{dumpcolor}{RGB}{255,140,0}
\definecolor{movecolor}{RGB}{50,180,50}
\vspace{-0.3em}

\begin{subfigure}[t]{0.62\linewidth}
\begin{subfigure}[t]{0.48\linewidth}
  \begin{tikzpicture}
    \node[anchor=south west,inner sep=0] (img) at (0,0) {\includegraphics[width=\linewidth]{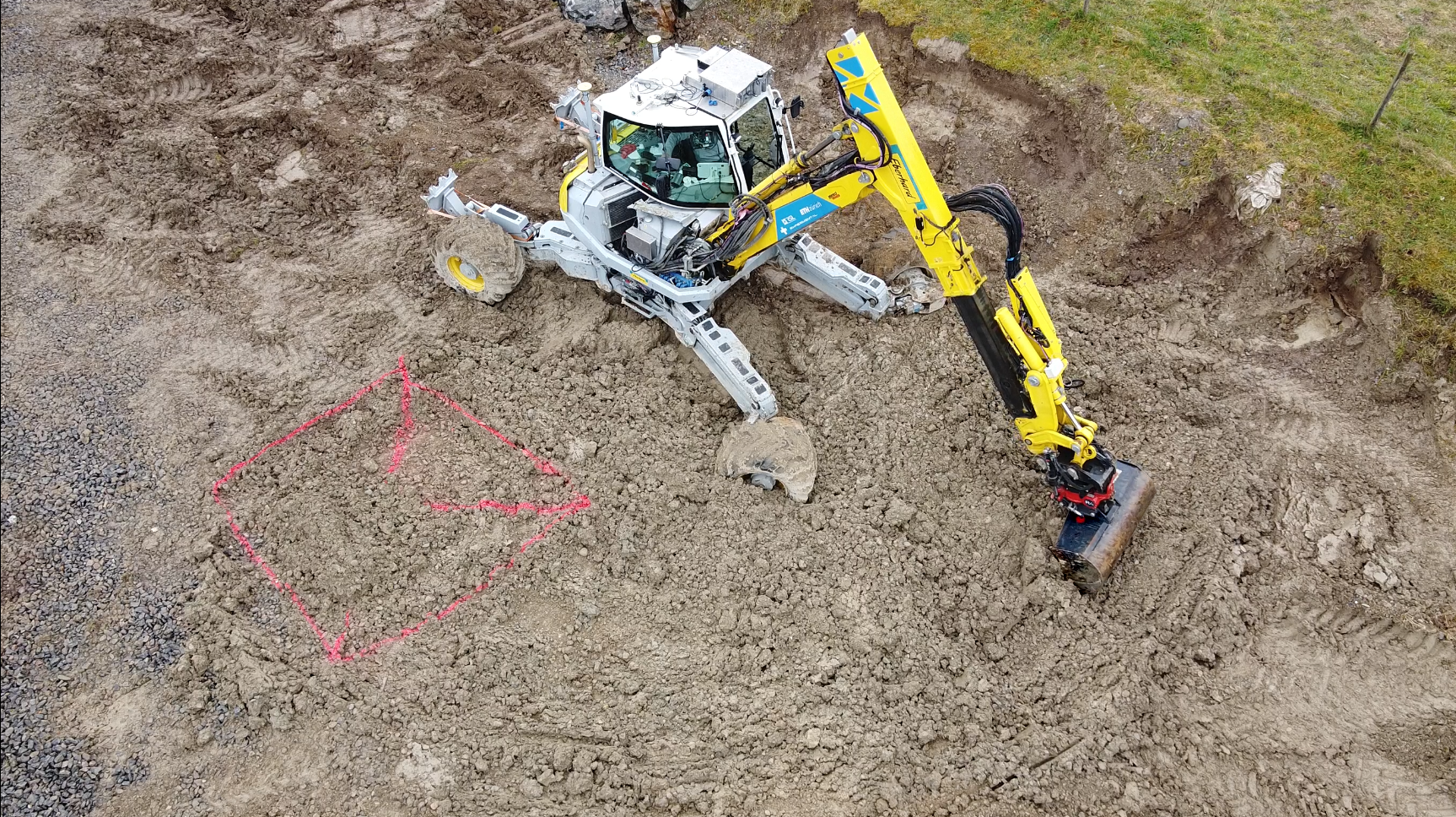}};
    \node[circle,fill=digcolor,text=white,font=\footnotesize,inner sep=1pt,minimum size=12pt] at (0.3,2.65) {$\mathbf{t_1}$};
  \end{tikzpicture}
  \\[-1.2em]
  \caption{\footnotesize Dig task (mid-scoop)}
  \label{fig:dig_execution_new}
\end{subfigure}%
\hfill%
\begin{subfigure}[t]{0.48\linewidth}
  \begin{tikzpicture}
    \node[anchor=south west,inner sep=0] (img) at (0,0) {\includegraphics[width=\linewidth]{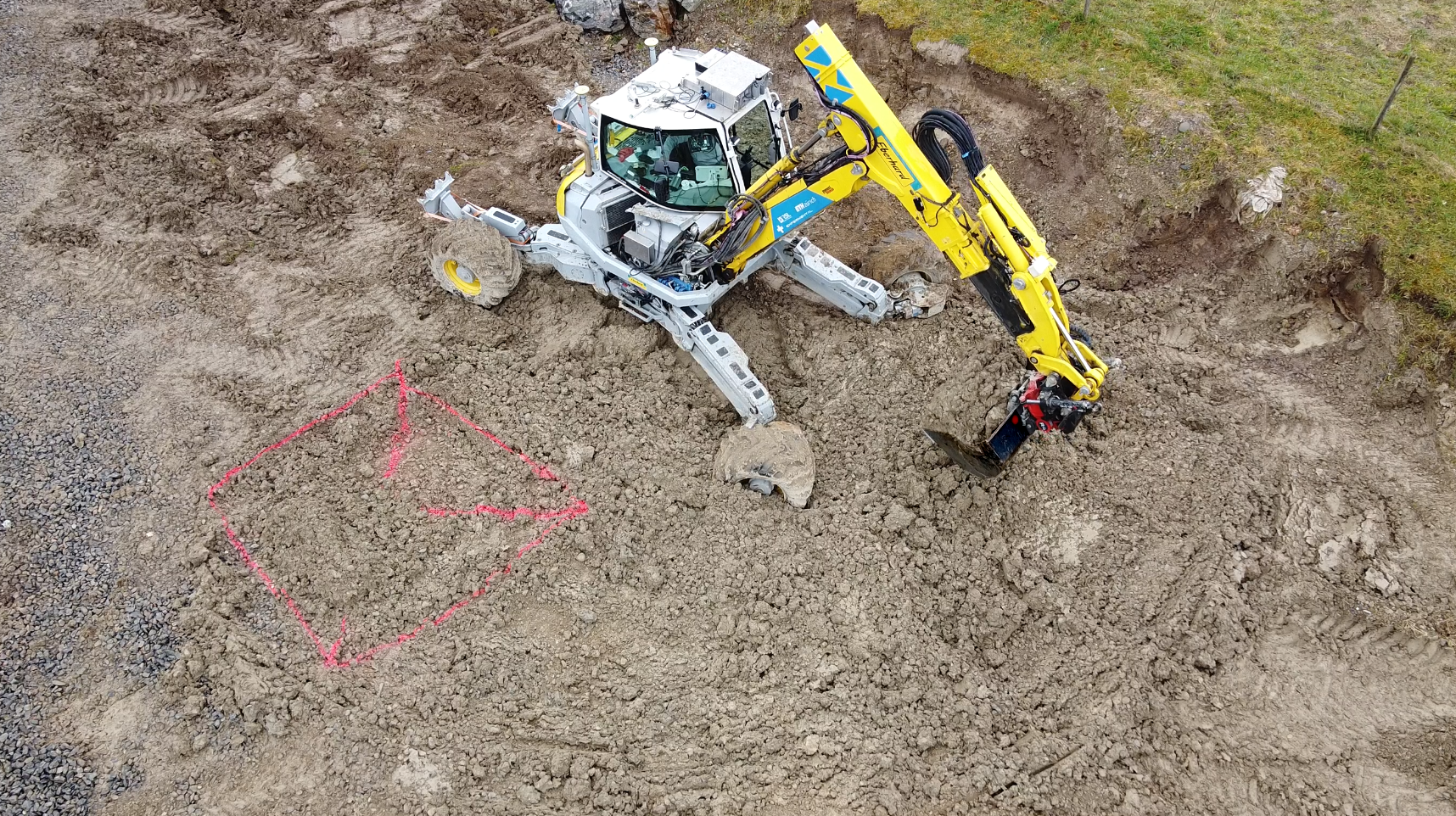}};
    \node[circle,fill=dumpcolor,text=white,font=\footnotesize,inner sep=1pt,minimum size=12pt] at (0.3,2.65) {$\mathbf{t_2}$};
  \end{tikzpicture}
  \\[-1.2em]
  \caption{\footnotesize Dig done, Dump begins}
  \label{fig:dig_to_dump_new}
\end{subfigure}

\vspace{0.05em}

\begin{subfigure}[t]{0.48\linewidth}
  \begin{tikzpicture}
    \node[anchor=south west,inner sep=0] (img) at (0,0) {\includegraphics[width=\linewidth]{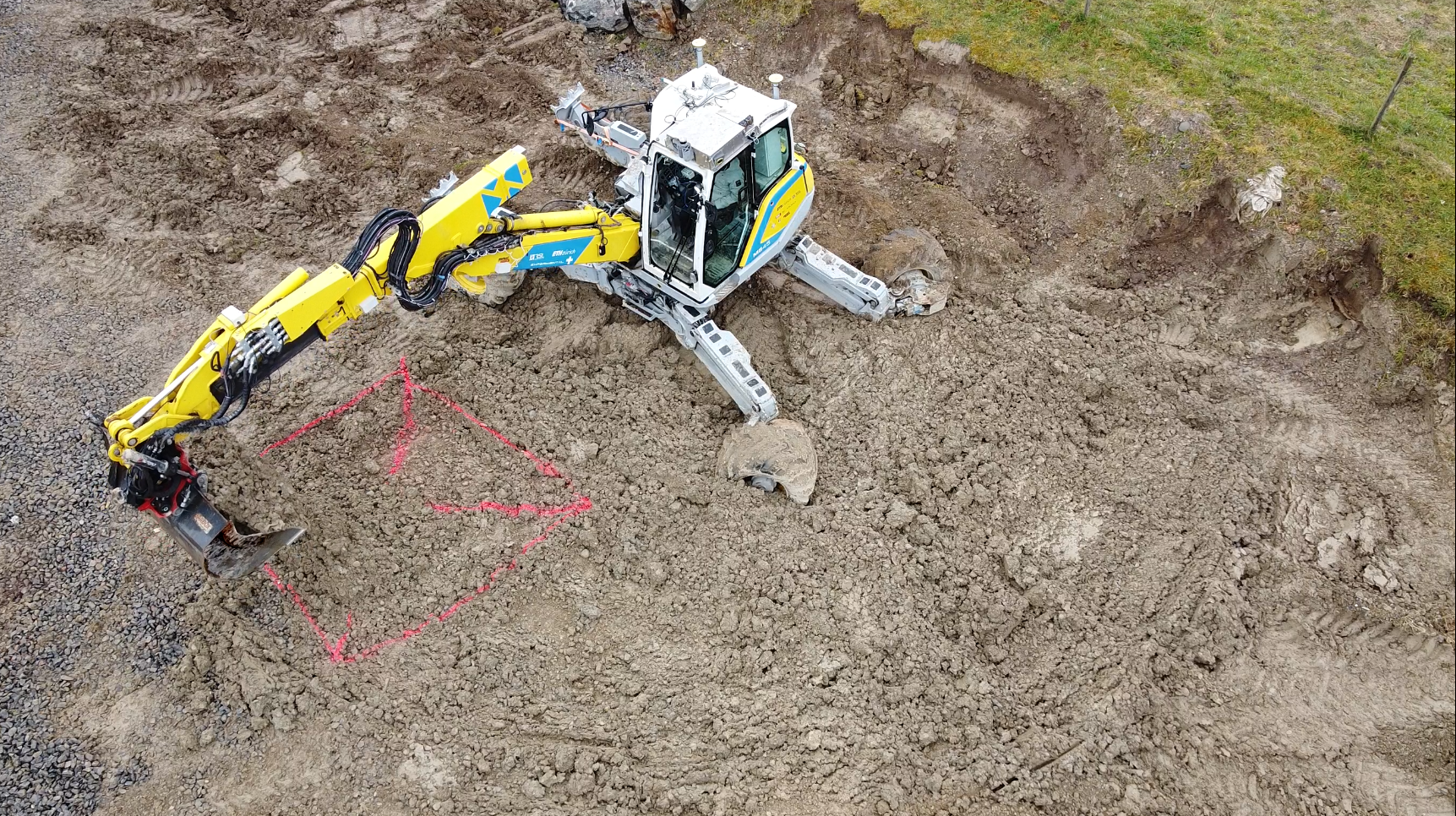}};
    \node[circle,fill=dumpcolor,text=white,font=\footnotesize,inner sep=1pt,minimum size=12pt] at (0.3,2.65) {$\mathbf{t_3}$};
  \end{tikzpicture}
  \\[-1.2em]
  \caption{\footnotesize Dump task (target reached)}
  \label{fig:dump_execution_new}
\end{subfigure}%
\hfill%
\begin{subfigure}[t]{0.48\linewidth}
  \begin{tikzpicture}
    \node[anchor=south west,inner sep=0] (img) at (0,0) {\includegraphics[width=\linewidth]{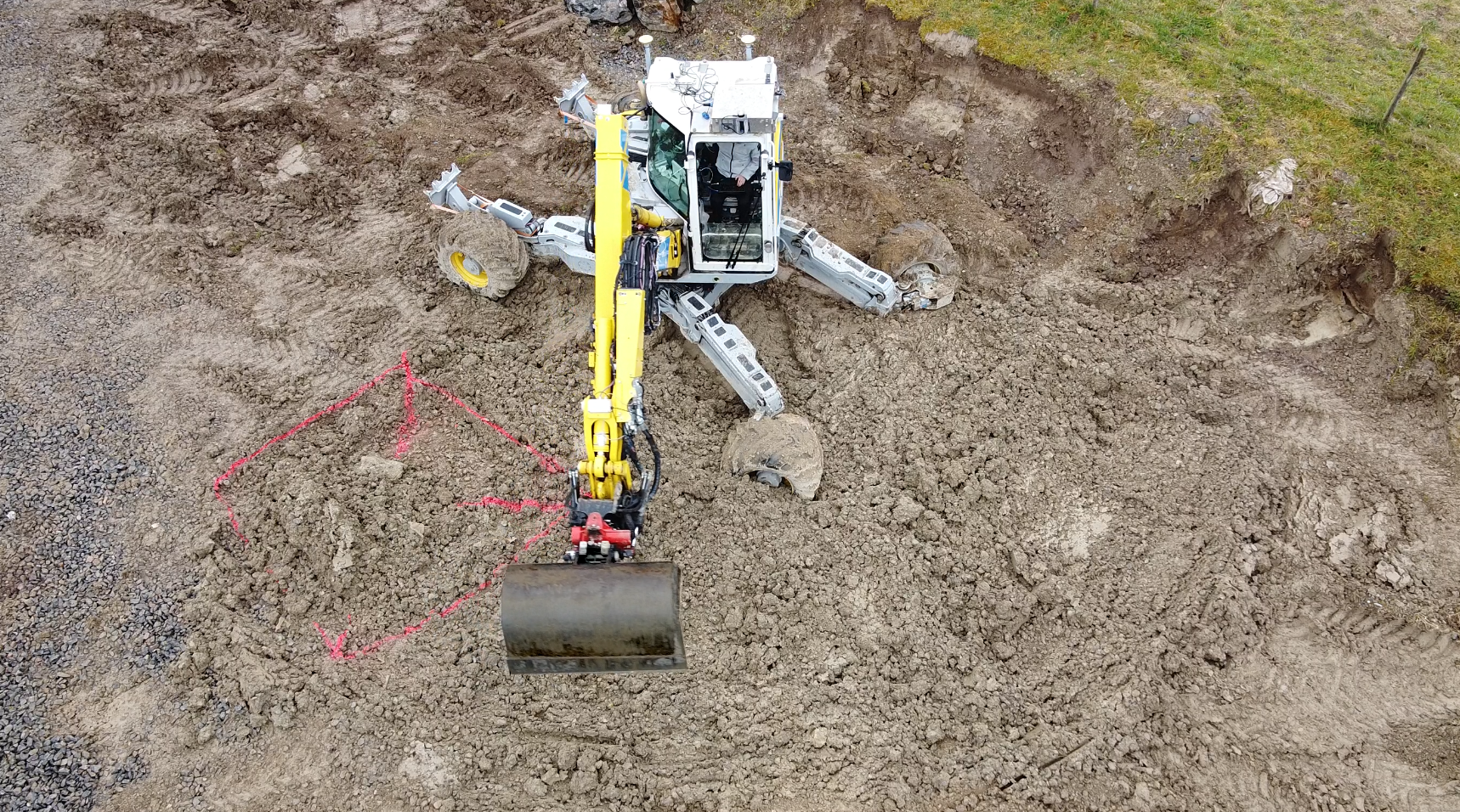}};
    \node[circle,fill=movecolor,text=white,font=\footnotesize,inner sep=1pt,minimum size=12pt] at (0.3,2.65) {$\mathbf{t_4}$};
  \end{tikzpicture}
  \\[-1.2em]
  \caption{\footnotesize Move Arm (repositioning)}
  \label{fig:move_execution_new}
\end{subfigure}

\vspace{-0.15em}
\begin{center}
\begin{tikzpicture}[scale=0.6,every node/.style={inner sep=0,outer sep=0}]
  \fill[digcolor] (0,0) rectangle (3,0.4);
  \fill[dumpcolor] (3,0) rectangle (7.5,0.4);
  \fill[movecolor] (7.5,0) rectangle (10.5,0.4);
  
  \node[font=\scriptsize,white] at (1.5,0.2) {$\mathbf{t_1}$};
  \node[font=\scriptsize,white] at (4.0,0.2) {$\mathbf{t_2}$};
  \node[font=\scriptsize,white] at (6.0,0.2) {$\mathbf{t_3}$};
  \node[font=\scriptsize,white] at (9,0.2) {$\mathbf{t_4}$};
  
  \node[font=\scriptsize] at (1.5,-0.2) {\textbf{Dig}};
  \node[font=\scriptsize] at (5.25,-0.2) {\textbf{Dump}};
  \node[font=\scriptsize] at (9,-0.2) {\textbf{Move}};
\end{tikzpicture}
\end{center}
\end{subfigure}%
\hfill%
\begin{subfigure}[t]{0.37\linewidth}
    \centering
    \raisebox{-0.16\height}{
        \begin{minipage}{\linewidth}
            \centering
            \includegraphics[width=0.93\linewidth, trim=100 0 140 0, clip]{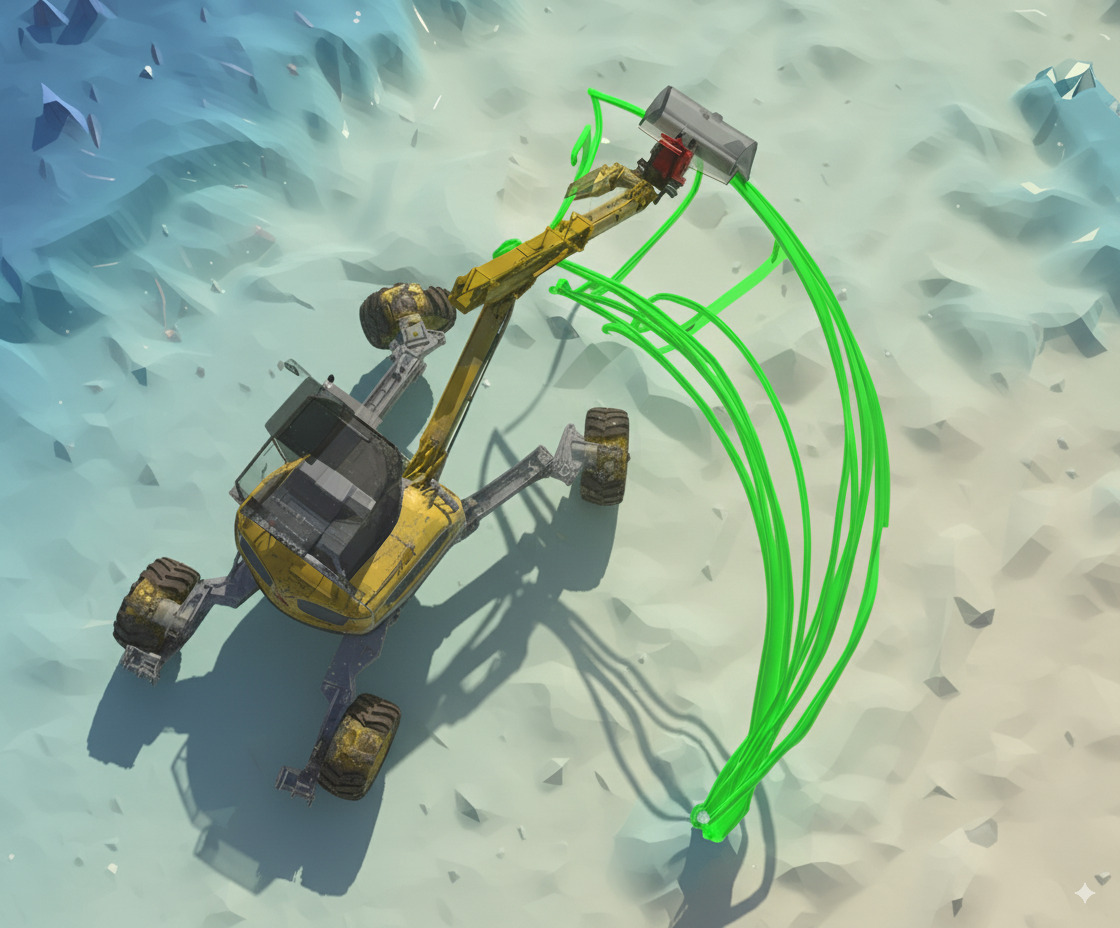}
            \vspace{-0.2em}
            \caption{Shovel trajectory visualization}
            \label{fig:workflow_and_validation}
        \end{minipage}
    }
\end{subfigure}

\vspace{-0.1em}
\caption{Complete autonomous excavation workflow using the pretrained GPT-4-Tasks policy on Menzi Muck M545: (a)-(d) real-world execution sequence with transitions between digging, dumping, and arm moving, (e) visualization of the end-effector trajectory throughout the six dig-dump-move cycles.}
\label{fig:excavation-workflow}
\end{figure*}
\begin{figure}[!hbt]
\centering
\definecolor{abortcolor}{RGB}{200,50,50}
\begin{subfigure}{0.47\linewidth}
  \begin{tikzpicture}
    \node[anchor=south west,inner sep=0] (img) at (0,0) {\includegraphics[width=\linewidth]{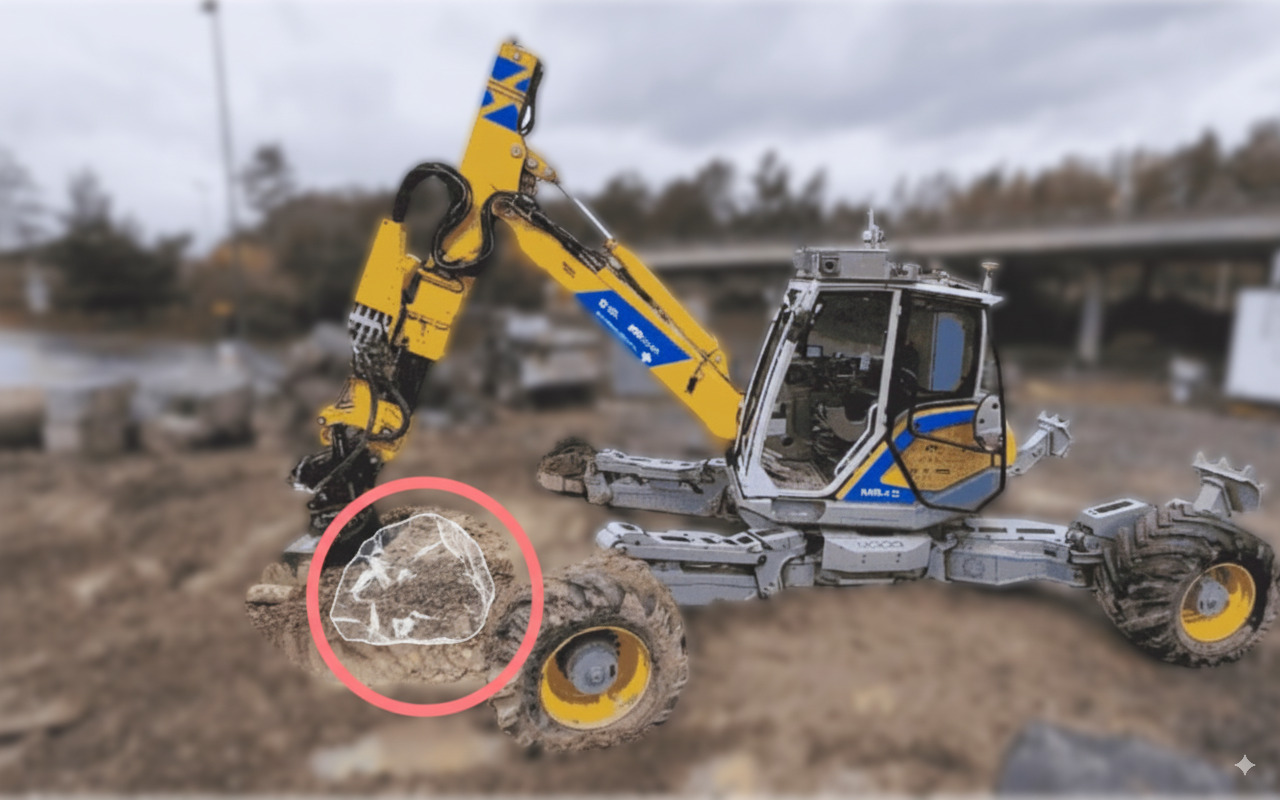}};
    \node[circle,fill=abortcolor,text=white,font=\footnotesize,inner sep=1pt,minimum size=12pt] at (0.3,2.2) {$\mathbf{t_0}$};
  \end{tikzpicture}
  \vspace{-0.5em}
  \caption{\footnotesize Collision with rock}
\end{subfigure}%
\hfill%
\hfill%
\begin{subfigure}{0.47\linewidth}
  \begin{tikzpicture}
    \node[anchor=south west,inner sep=0] (img) at (0,0) {\includegraphics[width=\linewidth]{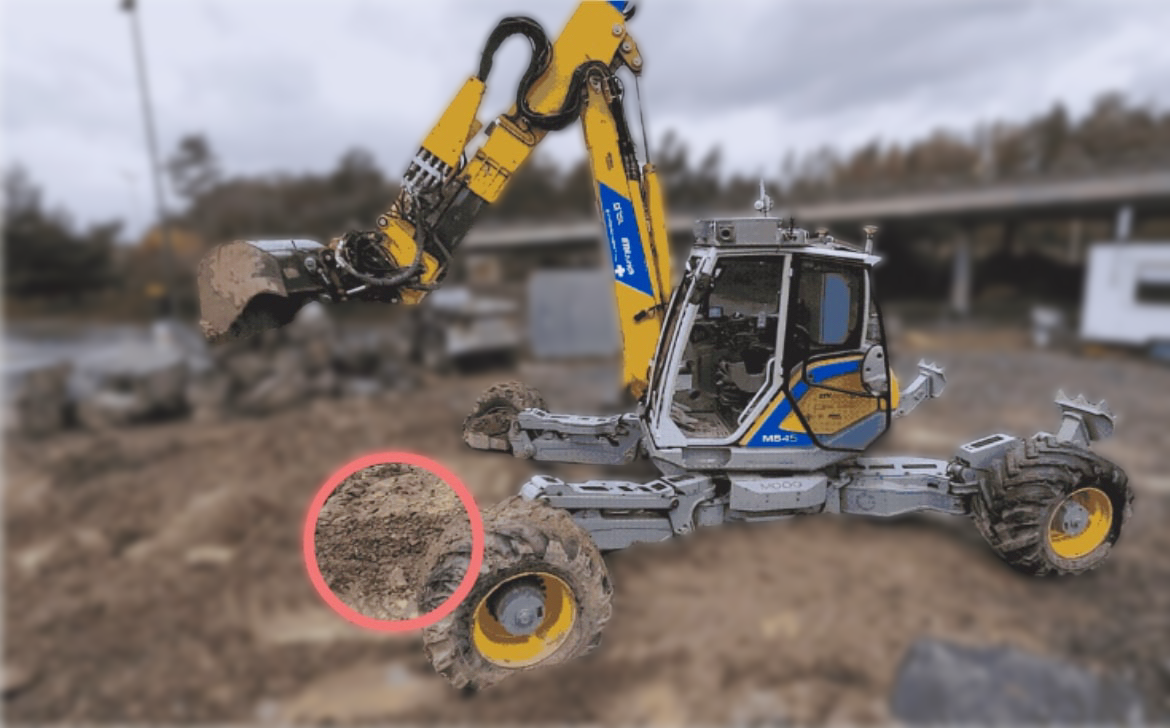}};
    \node[circle,fill=abortcolor,text=white,font=\footnotesize,inner sep=1pt,minimum size=12pt] at (0.3,2.2) {$\mathbf{t_1}$};
  \end{tikzpicture}
  \vspace{-0.5em}
  \caption{\footnotesize Recovery begins}
\end{subfigure}
\caption{Real-world collision recovery on the M545, with the buried rock obstacle circled in red, demonstrating the pretrained ExT policy's ability to detect stalls from proprioception and execute safe recovery maneuvers.}
\label{fig:obstacle-recovery}
\end{figure}
\begin{figure}[htb]
    \centering
    \includegraphics[trim=0 0 0 0, clip, width=0.8\linewidth]{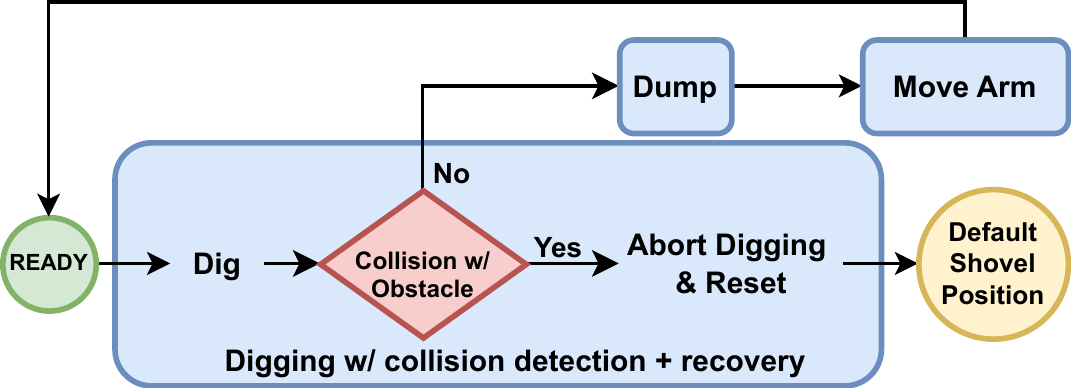}
    \caption{The Finite State Machine used to coordinate task switching during deployment of GPT-4-Tasks policy for a complete excavation workflow.}
    \label{fig:fsm}
\end{figure}

In preparation for deploying the pretrained models on the real machine, we first injected control delays (up to \SI{0.75}{\second}) and joint-velocity noise (uniformly sampled with an average amplitude of $\sim0.2 \;\mathrm{rad/s}$) in simulation to predict the robustness of ExT policies against potential sim-to-real gap from nonlinear hydraulic dynamics and imperfect velocity tracking. As shown in Table~\ref{sim_res_table}, the pretrained GPT policies remained robust to external disturbances for the \emph{dump} and \emph{move arm} tasks, and this robustness carried over to real-world deployment. Table~\ref{ik-results-real} summarizes the on-machine performance of these two tasks: the GPT policy pretrained across all four tasks successfully achieved zero-shot sim-to-real transfer, maintaining a mean error below \SI{4}{\cm}. We then deployed the GPT-4-Tasks policy in a complete excavation workflow coordinated by a simple finite state machine (Fig.~\ref{fig:fsm}). The system executed six dig-dump-move cycles at three designated dig points and a target dumping location \SI{90}{\degree} clockwise from the forward direction, integrating all four tasks seamlessly (Fig.~\ref{fig:excavation-workflow}, Fig.~\ref{fig:obstacle-recovery}). The system achieved an average excavation volume of \SI{0.96}{\cubic\meter} per cycle (\SI{0.68}{\cubic\meter} bucket capacity) and an average cycle time of \SI{36}{\second}, comparable to prior methods \cite{towards_auto_ex}. The task-level accuracy was also consistent, with real digging locations $10.7 \pm 3.0$ \si{\centi\meter} and dumping locations $6.5\pm2.9$ \si{\centi\meter} away from the specified targets. These results highlight the real-world applicability of policies trained with the ExT framework.

\subsection{Few-Shot Skill Acquisition via SFT}
\label{sec:exp_sft}

With \ac{sft}, we explore rapid skill acquisition from limited demonstrations. We start with a GPT model pretrained on the \emph{dig} and \emph{move arm} datasets (150k demonstrations each) and fine-tune it to acquire the \emph{dump} skill using only 1k demonstrations—an amount feasible for human data collection. To mitigate catastrophic forgetting, a small replay buffer of $500$ \emph{dig} and $350$ \emph{move arm} episodes is interleaved during fine-tuning. Table~\ref{tab:few_shot} compares this \ac{sft} approach with two baselines: a GPT model trained from scratch on the same 1k demonstrations, and another trained from scratch on the full 150k dataset. The pretrained+SFT model achieves high success rates on \emph{dump} while maintaining strong performance on the original tasks, despite using two orders of magnitude less data than full pretraining. Although models pretrained on the full \emph{dump} dataset remain more performant overall (Table~\ref{sim_res_table}), few-shot \ac{sft} recovers a large portion of this performance at a fraction of the data cost. In contrast, training from scratch on only 1k demonstrations fails to learn the task, showing failure modes such as off-track motion, premature bucket opening, and overshoot near the target. These results underscore the advantage of pretrained policies on efficient skill acquisition in data-scarce regimes.

\begin{table}[t]
\centering
\caption{\ac{sft} results: The pretrained model (trained on \emph{dig} and \emph{move arm} demos) acquires the \emph{Dump} task using 150x less data than a single-task model trained from scratch, achieving comparable performance while preserving high accuracy on the original pretraining tasks.$^{\dagger}$}
\label{tab:few_shot}
\setlength\tabcolsep{5pt}
\begin{tabular}{lccc}
\toprule
& Training & \multicolumn{2}{c}{\emph{Dump} Task Performance} \\
\cmidrule(lr){3-4}
Method & Data & Success & Error (cm) \\
\midrule
GPT Pretrained+SFT & 1k   & 96\%  & 4.3 $\pm$ 3.0 \\
GPT BC from scratch   & 1k   & 12\%  & 45.7 $\pm$ 30.3 \\
GPT BC from scratch & 150k & 100\% & 2.4 $\pm$ 0.9 \\
\bottomrule
\multicolumn{4}{@{}l}{\footnotesize $^{\dagger}$Performance on pretrained tasks: 98\% \emph{dig} success,  \SI{1.5}{\cm} \emph{move arm} error.}
\end{tabular}
\end{table}

\subsection{Reinforcement Learning Fine-Tuning}
\label{sec:rlft}

\begin{figure}[t]
  \centering
  \includegraphics[width=0.48\linewidth, trim={80 200 80 180}, clip]{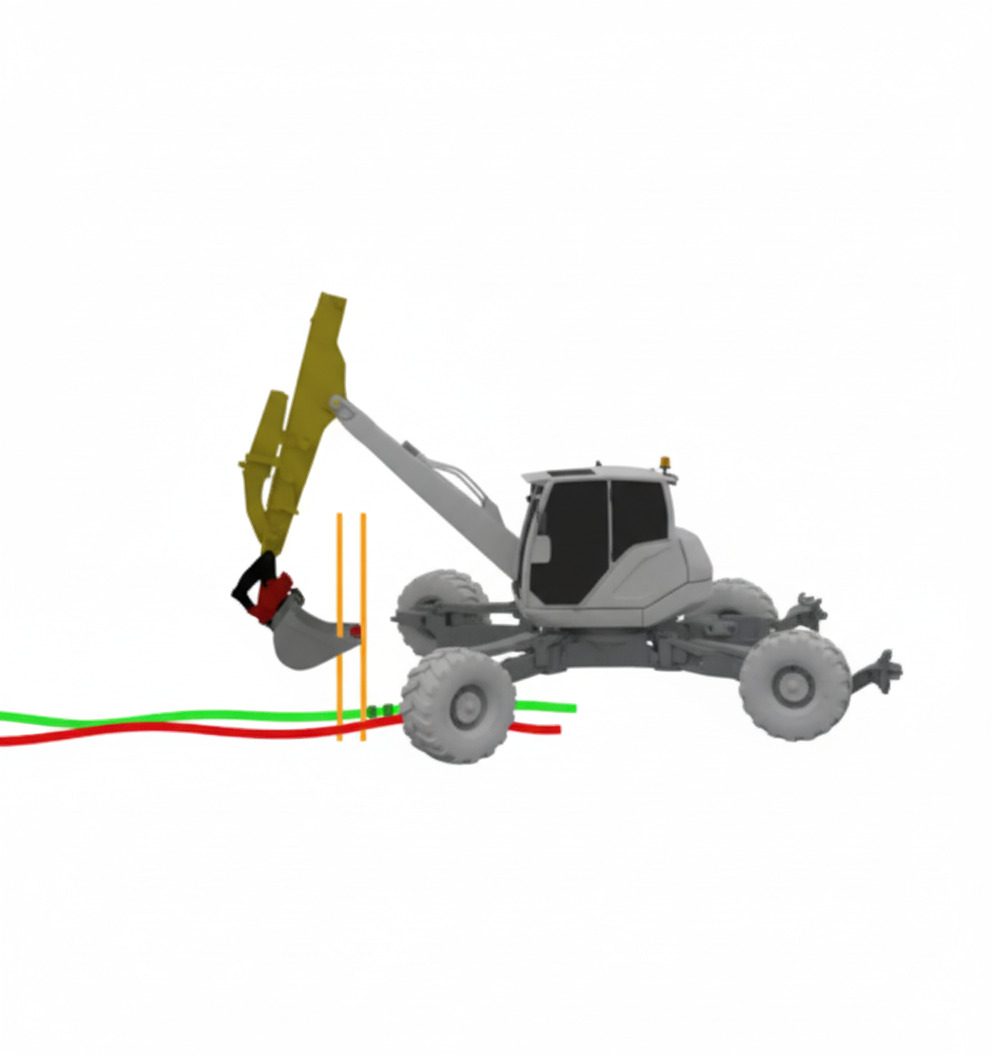}
  \hfill
  \includegraphics[width=0.48\linewidth, trim={150 280 150 220}, clip]{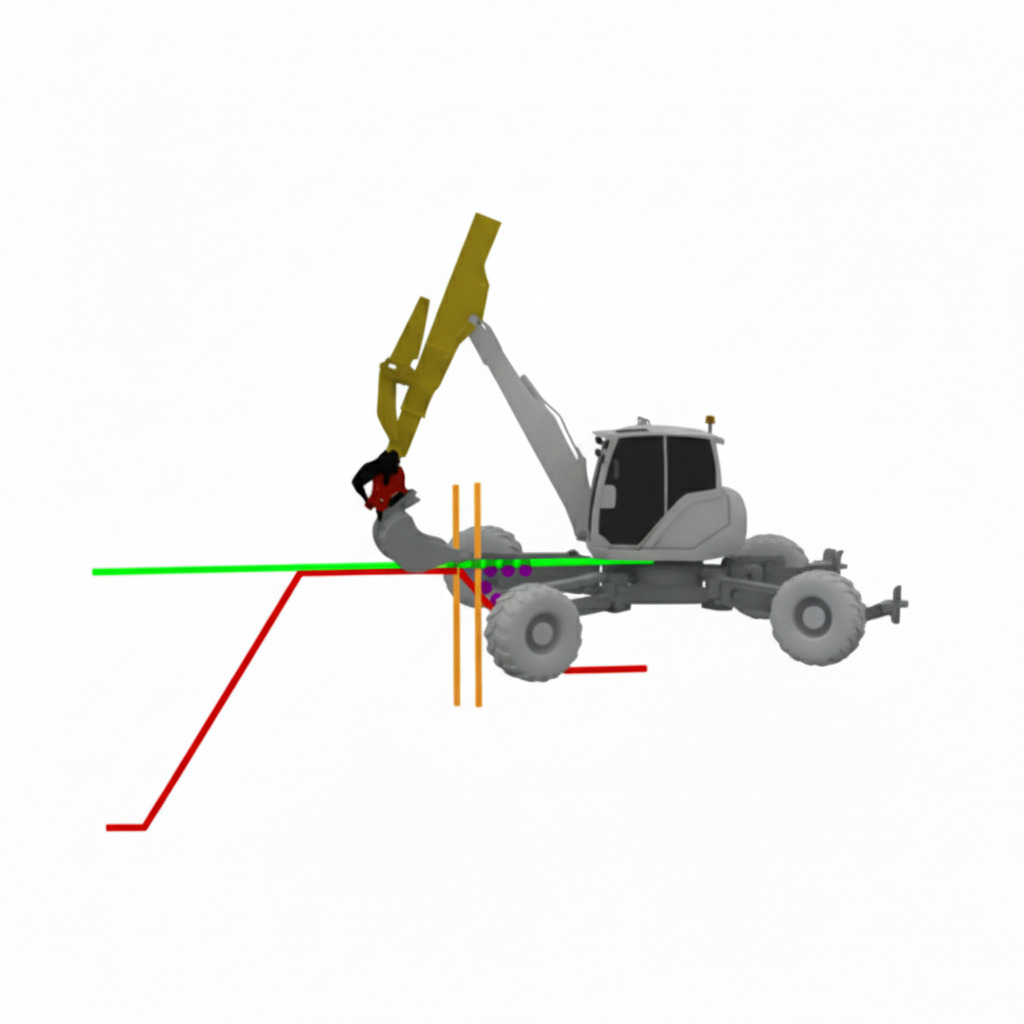}
  \caption{Change in soil terrain (max digging depth profile in red) from RBF-sampled terrain (left) to stair-shaped terrain (right).}
  \label{fig:soil_shapes}
\end{figure}

Using \ac{rlft}, we aim to adapt a pretrained policy to out-of-distribution (OOD) conditions for the \emph{dig} task. We consider three types of distribution shifts: (i) a change in the maximum digging depth profile from a radial basis function (RBF) to a stair-shaped terrain, visualized in Fig.~\ref{fig:soil_shapes}, (ii) OOD soil parameters in the fundamental earth moving equations used for soil simulation \cite{pascal-new-rl}, and (iii) a new excavator shovel geometry. The specific parameter changes are summarized in Table \ref{tab:soil-params}. These shifts reflect common unseen conditions encountered when deploying autonomous excavation systems at worksites, including new excavation soil profile requirements, previously unseen soil types, and new excavator attachments.

\begin{table}[t]
\begin{center}
\caption{Parameter ranges used for \emph{dig} demonstration collection, along with the out-of-distribution values to be adapted by the policy with \ac{rlft}.}
\label{tab:soil-params}
\begin{tabular}{l c c c | c}
\toprule
Parameters & Unit & Min & Max & OOD Value\\
\midrule
\myrowcolour
Terrain RBF length scale & m & 0 & 0.5 & Stairs\\
\midrule
Cohesion & kPa & 0 & 100 & 0\\
\myrowcolour
Adhesion & --- & 0 & 1 & 0\\
Soil internal friction angle & rad & 0.3 & 0.8 & 0.77 \\
\myrowcolour
Soil--bucket friction angle & rad & 0.19 & 0.38 & 0.4 \\
Cavity pressure factor & --- & 0 & 300 & 315\\
\midrule
\myrowcolour
Bucket width & m & 1.4 & --- & 1.1\\
Bucket radius & m & 0.375 & --- & 0.275\\
\bottomrule
\end{tabular}
\end{center}
\end{table}

\begin{table}[t]
\begin{center}
\caption{Training Configuration for \ac{rlft}}
\label{tab:ppo-hparams}
\begin{tabular}{ l c }
\toprule
\multicolumn{1}{l}{Hyperparameter} & \multicolumn{1}{c}{Value} \\
\midrule
\myrowcolour
Number of environments & 1000 \\
Steps between PPO iterations & 6 \\
\myrowcolour
Number of PPO iterations & 100 \\
Mini-batches per PPO iteration & 10 \\
\myrowcolour
Total environment interactions & 600,000 \\
Learning rate (LR) & $10^{-5}$ \\
\myrowcolour
Minimum LR (cosine annealing) & $10^{-7}$ \\
\bottomrule
\end{tabular}
\end{center}
\vspace{-0.3cm}
\end{table}

\begin{figure}[!t]
\centering
\includegraphics[width=\columnwidth,  trim={6 0 0 0}, clip]{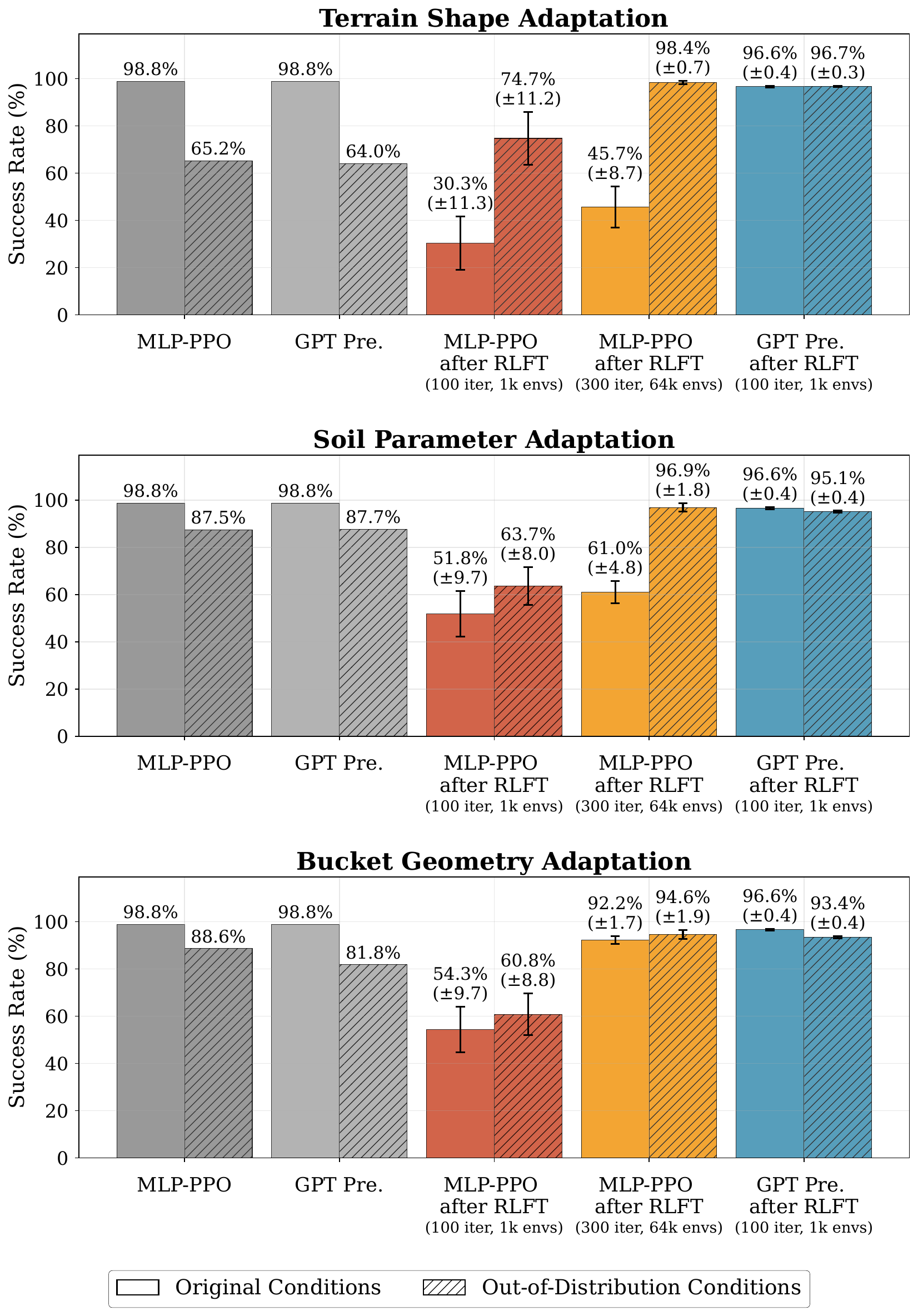}  
\caption{RLFT adaptation performance across three distribution shifts. Pretrained GPT policy adapts efficiently with minimum compute (100 iterations, 1k environments) while maintaining high success on original conditions. The MLP-PPO policy requires significantly more data (300 iterations, 64k environments) to match adaptation performance.}
\label{fig:rlft_adaptation}
\end{figure}

We conduct \ac{rlft} experiments in simulation on a GPT model pretrained on \emph{dig}, \emph{dump}, and \emph{move arm} tasks to evaluate adaptation to these distribution shifts in the \emph{dig} environment. As a baseline comparison, we also apply \ac{rlft} on an MLP policy trained from scratch with \ac{ppo} on the original \emph{dig} task settings, which is also the \ac{rl} expert used to collect \emph{dig} demonstrations. All policies are trained and evaluated over five random seeds using the configuration in Table~\ref{tab:ppo-hparams}, which features limited parallel environments, batch size, and total environment interactions. Reward and termination conditions follow the implementation in \cite{pascal-new-rl}. We report the mean success rates with $95\%$ confidence intervals. Results are summarized in Fig. \ref{fig:rlft_adaptation}.

Without \ac{rlft}, both MLP-PPO and the pretrained GPT model experience performance drops under OOD conditions. Interestingly, their performances in these conditions are highly similar, suggesting successful distillation of the MLP-PPO expert into the GPT model through large-scale pretraining. After applying \ac{rlft}, despite limited environment interactions, the GPT fine-tuned model achieves high success rates on the OOD conditions while showing minimal forgetting on the original task configurations. In contrast, applying the same fine-tuning schedule to the MLP-PPO expert policy leads to poor performance. This can be attributed to the small batch size, which increases variance in advantage estimation and results in noisy policy updates. Such updates are insufficient for effective adaptation in the MLP architecture, causing significant degradation on both the original and OOD conditions. On the other hand, the robust representations and behavior priors learned by the GPT model during pretraining mitigate this issue, enabling effective learning in this low-data regime.

To test whether this performance gap can be closed, we increase the training budget for the MLP-PPO expert to 300 iterations with $64\text{k}$ parallel environments, totaling $115.2M$ environment interactions. Under this larger-scale setting, the MLP-PPO policy eventually achieves strong performance on the OOD tasks but continues to suffer from substantial forgetting on the original configurations. This contrast highlights two key advantages of the pretrained GPT model: (i)~superior sample efficiency and (ii)~stronger retention of prior knowledge during adaptation.


\begin{figure}[!t]
\centering
\includegraphics[width=\columnwidth, trim={0 0 0 0}, clip]{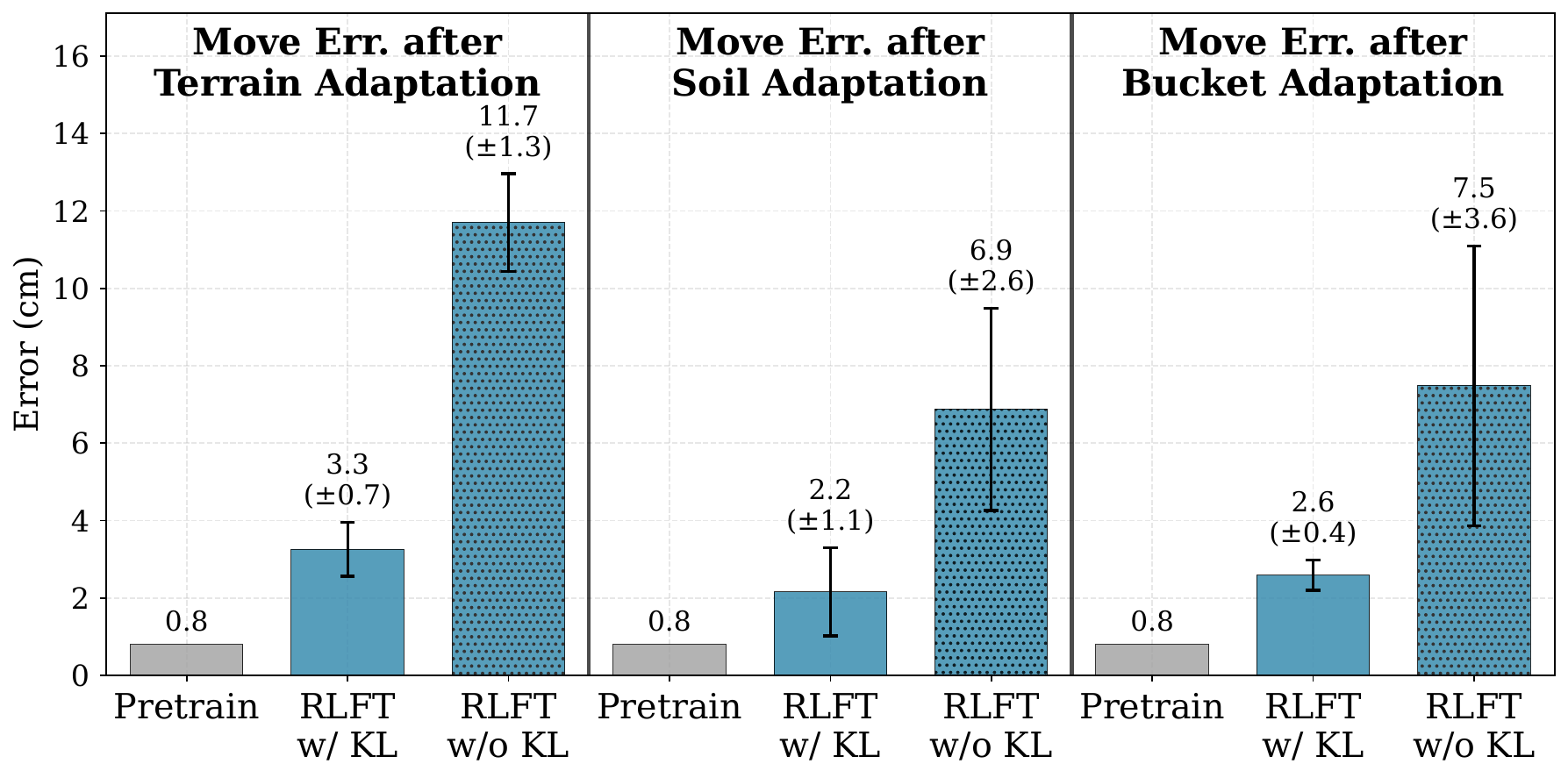}
\caption{The effect of KL regularization on pretrained task performance retention: \emph{move arm} end-effector position error after RLFT, with and without KL regularization.}
\label{fig:precision_errors}
\end{figure}

Finally, we perform an ablation study to evaluate the effect of KL regularization on retaining pretrained task performance. Specifically, we apply \ac{rlft} both with and without the KL penalty term to the pretrained GPT policy and compare the fine-tuned policies' performance on the \emph{move arm} task. As shown in Fig.~\ref{fig:precision_errors}, removing the KL term leads to substantial degradation of the \emph{move arm} skill across all three adaptation scenarios, evidenced by higher end-effector position errors. In contrast, incorporating the KL penalty into the \ac{rlft} objective effectively constrains excessive deviation from the pretrained prior, thus reducing forgetting. Importantly, we observe that the KL regularization term does not affect the performance of the fine-tuned policy on the OOD conditions, with a difference of $-0.6\%$, $-0.4\%$, $+1.3\%$ relative to the fine-tuned policy trained without KL regularization. These results demonstrate that KL regularization provides a simple yet effective mechanism to preserve pretrained skills while enabling robust adaptation to new tasks.


\section{Conclusion}

We introduce ExT, a unified framework for large-scale multi-task demonstration generation, pretraining, and fine-tuning of excavator policies. Through both simulation and on-machine experiments, we demonstrate that pretrained ExT policies can successfully perform a complete excavation workflow zero-shot in a sim-to-real setting. This capability stems from the robustness and versatility of the large-scale, multi-task transformer policy trained on high-quality demonstrations generated by ExT. We validate ExT's fine-tuning pathways: \ac{sft} for tasks with limited available demonstrations, and \ac{rlft} for operating conditions where reward functions can be modeled in simulation. Using these fine-tuning techniques, new skills can be acquired rapidly with few demonstrations, and out-of-distribution conditions can be adapted efficiently in simulation. These results highlight ExT's potential to enable scalable and adaptable deployment of excavation policies in real-world scenarios. Future work for ExT includes expanding the framework to support a wider range of excavator models and tasks, enhancing the policy's versatility for scalable autonomous excavation. Integrating unified perception using both LiDAR and vision will enable end-to-end learning of policies, improving decision-making in complex environments. Furthermore, extending the framework to encompass soil manipulation tasks, leveraging GPU-accelerated MPM simulations, will allow for more accurate and realistic modelling of soil dynamics, further closing the sim-to-real gap and enable more complex manipulation tasks.

\bibliographystyle{IEEEtran}
\bibliography{bib}
\end{document}